\documentclass[journal]{IEEEtran}

\usepackage{xcolor,soul,framed} 

\colorlet{shadecolor}{yellow}
\usepackage[pdftex]{graphicx}
\graphicspath{{../pdf/}{../jpeg/}}
\DeclareGraphicsExtensions{.pdf,.jpeg,.png}

\usepackage{array}
\usepackage{eqparbox}
\usepackage{url}

\usepackage[cmex10]{amsmath}
\usepackage{amssymb}
\usepackage{amsthm}

\usepackage{xcolor} 

\usepackage{doi} 
\usepackage{cleveref}
\usepackage{booktabs}
\usepackage{multirow}

\newtheorem{definition}{Definition} 

\definecolor{red2}{HTML}{f20c00}
\definecolor{green2}{HTML}{19a951}
\definecolor{grayc6}{HTML}{DAE8FC}
\definecolor{myblue}{HTML}{045AAE}

\usepackage{algorithm}      
\usepackage{algpseudocode}  
\usepackage{caption}        
\usepackage{subcaption} 
\usepackage{diagbox}

\usepackage{mdwmath}
\usepackage{mdwtab}

\begin{document}
\bstctlcite{IEEEexample:BSTcontrol}
    \title{Robust Fuzzy Multi-view Learning under View Conflict}
  \author{Siyuan Duan,
      Yuan Sun,
      Dezhong Peng,
      Yingke Chen,
      Xi Peng,
      Peng Hu$^*$

  \thanks{Siyuan Duan, Yuan Sun, and Peng Hu are with the College of Computer Science, Sichuan University, Chengdu 610065, China.}
  \thanks{Dezhong Peng is with the College of Computer Science, Sichuan University, Chengdu 610065, China, and also with the Tianfu Jincheng Laboratory, Chengdu 611130, China.}
  \thanks{Yingke Chen is with the Department of Computer and Information Sciences, Northumbria University, Newcastle upon Tyne NE1 8ST, UK}
  \thanks{Xi Peng is with the School of Artificial Intelligence, Sichuan University, Chengdu 610000, China.}
  \thanks{Corresponding author: Peng Hu. (email: penghu.ml@gmail.com).}}


\maketitle




\begin{abstract}
Trusted multi-view classification aims to deliver reliable fusion for accurate predictions and has recently attracted substantial attention in both academia and industry. However, existing TMVC methods typically assume strict alignment across different views during both training and testing phases, which is often impractical in real-world scenarios. This limitation motivates us to revisit TMVC and extend it to a more challenging setting: \textit{how to mitigate the impact of view conflict (VC) during both training and inference}. To tackle this setting, existing TMVC methods suffer from three critical limitations: underestimated uncertainty, misleading decisions, and overfitting to VC. To address these issues, this paper proposes a novel \underline{R}obust \underline{Fu}zzy \underline{M}ulti-View \underline{L}earning (R-FUML) framework grounded in Fuzzy Set Theory. Specifically, R-FUML models network outputs as fuzzy memberships to quantify category credibility and uses an entropy-based method for reliable multi-view fusion. To this end, we present a Robust Multi-view Fusion (RMF) strategy that accounts for both view-specific uncertainty and inter-view conflicts, thereby alleviating the adverse impacts of VC on decision-making. To identify and conquer VC during training, we further design a Robust Learning Against VC (RLVC) framework. RLVC isolates conflicting samples by leveraging neural networks' memory effects and then retrains the model by applying a penalty to these conflicting views. Extensive experiments across eight public datasets demonstrate that R-FUML consistently outperforms 15 state-of-the-art baselines in robustness and uncertainty estimation. The code will be released upon acceptance.
\end{abstract}




%
\IEEEpeerreviewmaketitle


\section{Introduction}

\begin{figure}[tbp]
    \centering
    \includegraphics[width=0.44\textwidth]{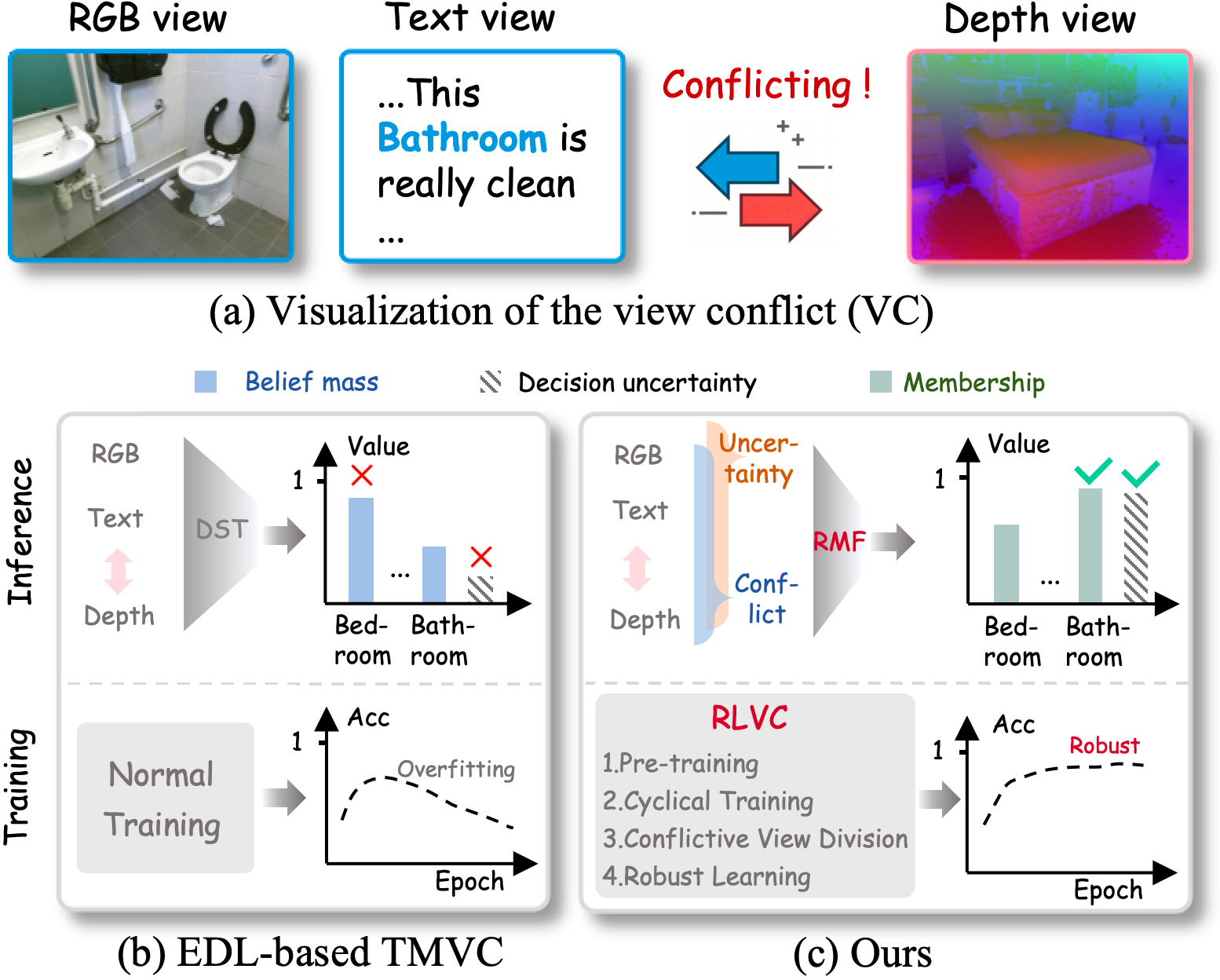}
    \caption{Illustration of view conflict (VC) and the motivation of our R-FUML. (a) Example of VC: the RGB and text views are both consistent with the `Bathroom' category, whereas the Depth view contains conflicting information, e.g., `Bedroom'. (b) Existing EDL-based TMVC methods rely on DST for fusion, which is often misled by VC and yields unreliable uncertainty estimates. Conventional training may also cause neural networks to overfit VC. (c) R-FUML addresses VC through two components: RMF mitigates its effect during inference, and RLVC improves robustness during training.}
    \label{toyExample}
    \vspace{-0.5cm}
\end{figure}

\IEEEPARstart{M}{ulti}-view/modal data, such as text, images, videos, and 3D point clouds~\cite{guo2020deep,feng2026robust}, have become ubiquitous in the modern world~\cite{zhang2024multimodal}. By exploiting complementary information across different views, multi-view learning enables capabilities beyond single-view learning and has been widely applied to video surveillance~\cite{wang2024real}, autonomous driving~\cite{gong2025carp}, sentiment analysis~\cite{yadav2023deep}, and medical diagnosis~\cite{xu2025distilled}. 
However, reliable use of such multi-source information remains fundamentally challenging. In safety-critical and practical scenarios, multi-view classification systems must provide not only accurate predictions but also accurate uncertainty estimates, which are essential indicators of prediction credibility and trustworthiness. Such uncertainty estimates support risk-aware decision-making by providing actionable guidance, e.g., in medical applications, they can help clinicians decide whether to rely on model outputs or pursue additional screening.

To enable reliable decision-making, a series of trusted multi-view classification methods (TMVC) have been proposed~\cite{han2022trusted}. These methods produce accurate classification results while estimating decision uncertainty, i.e., the confidence level associated with a prediction. Existing TMVC methods usually incorporate Evidential Deep Learning (EDL)~\cite{sensoy2018evidential} into multi-view learning architectures and utilize Dempster-Shafer theory (DST)~\cite{shafer1992dempster} to fuse view-specific evidence for uncertainty quantification. Along this line, some studies focus on improving evidence aggregation strategies to enhance performance~\cite{liu2022trusted,liu2024dynamic,liu2025enhancing}, whereas others extend TMVC to more realistic scenarios, such as learning with incomplete views~\cite{xie2023exploring}, label noise~\cite{ijcai2024p582}, and view adversarial attacks~\cite{wang2025reliable}. Despite this progress, existing EDL-based TMVC methods usually implicitly assume strict alignment across different views during both training and testing.

In practice, this assumption is often unrealistic or even infeasible because of sensor failures, adverse weather conditions, data communication issues, and other inevitable factors during both data collection and model inference~\cite{huang2024noise,lai2025rethinking}. Specifically, such factors can induce view conflict (VC), as illustrated in~\Cref{toyExample}(a), in both the training and test sets. This gap motivates us to revisit TMVC and reveal a more challenging problem: 
\textit{how to mitigate the impact of view conflict (VC) during both training and testing.}
However, existing EDL-based TMVC methods lack effective mechanisms to tackle this challenge, leading to three critical issues:
i) their uncertainty estimation neglects conflicts among category predictions, leading to underestimated uncertainty when the number of categories is small relative to the total evidence and consequently yielding inaccurate uncertainty quantification for VC-affected multi-view instances; 
ii) their multi-view fusion strategy based on Dempster-Shafer theory (DST)~\cite{shafer1992dempster}   fails to account for global conflicts among views and tends to overemphasize dominant evidence~\cite{xiao2019multi,shang2021compound}, often leading to misclassification for instances with VC; 
iii) they tend to overfit instances with VC, resulting in performance degradation, as shown in~\Cref{toyExample}(b).

\begin{figure*}[tbp]
    \centering
    \includegraphics[width=0.9\textwidth]{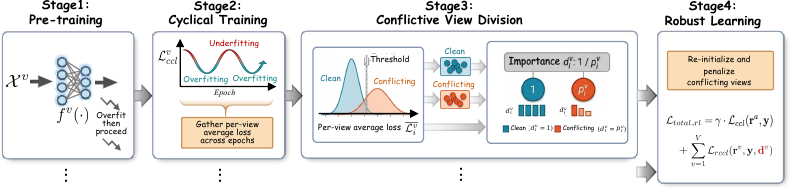}
    \caption{
    Overview of the Robust Learning Against VC (RLVC) framework. Stage 1 trains the network to overfitting. Stage 2 cyclically alternates between overfitting and underfitting. Stage 3 records per-sample loss trajectories to adaptively partition clean and conflicting samples. Finally, Stage 4 re-initializes and retrains the network while penalizing conflicting views.
    }
    \label{method2}
    \vspace{-0.5cm}
\end{figure*}

To address these issues, this paper proposes a novel framework, \underline{R}obust \underline{Fu}zzy \underline{M}ulti-View \underline{L}earning (\textbf{R-FUML}). Unlike EDL-based methods, our R-FUML is grounded in Fuzzy Set Theory~\cite{zadeh1965fuzzy}, which models inherent fuzziness by introducing membership degrees in $[0, 1]$. This formulation allows samples to belong to multiple categories with different degrees, thereby supporting effective uncertainty quantification. 
Building on this principle, \textbf{first}, we model network outputs as category-specific memberships. However, memberships provide only a possibility measure and do not reflect the extent to which a sample does not belong to other categories. To address this limitation, we introduce category credibility. We further propose an entropy-based uncertainty estimation method that utilizes category credibility to enhance uncertainty quantification performance.
\textbf{Second}, to improve classification accuracy for multi-view instances with VC, we propose Robust Multi-view Fusion (RMF), which downweights high-uncertainty and conflicting views, as illustrated in~\Cref{method1}.
\textbf{Finally}, to reduce the impact of VC during training, we propose a Robust Learning Against VC (RLVC) framework, which consists of four stages (see in~\Cref{method2}). In the `pre-training stage', the network is trained until overfitting. 
In the `cyclical training stage', we cyclically shift the model state between overfitting and underfitting and record the loss of each sample across all epochs. In the `conflicting view division stage', we employ an adaptive method to separate clean and conflicting samples. 
Finally, in the `robust learning stage', we re-initialize the network parameters and re-train the model by penalizing the conflicting views with a penalty term. 
By combining fuzzy-based modeling with RMF and RLVC, our R-FUML achieves better robustness and more accurate uncertainty estimation for multi-view instances with VC, as shown in~\Cref{toyExample}(c). 
In summary, the main contributions of this paper are as follows:
\begin{itemize}
    \item We reveal and explore a challenging TMVC problem, in which the model must be trainable on multi-view instances with VC while still supporting accurate classification. To the best of our knowledge, this work could be the first attempt to explore this problem.
    \item To address this problem, we propose the \textbf{R-FUML} method grounded in Fuzzy Set Theory, which enables more accurate uncertainty quantification through category-specific membership and credibility modeling.
    \item Furthermore, R-FUML incorporates two key technical components: RMF, which downweights high-uncertainty and conflicting views for reliable fusion, and RLVC, which enables robust training by identifying and penalizing conflicting samples.
    \item We conduct extensive experiments on eight public datasets and compare our R-FUML with 15 state-of-the-art baselines, demonstrating the superiority of the proposed method in robustness and uncertainty estimation.
\end{itemize}

\section{Related work}
\subsection{Multi-view Learning}

Multi-view learning (MVL)~\cite{liu2025reliable, wen2026multi} has been widely studied to exploit cross-view complementarity to boost representation quality and predictive performance. Classical methods include Canonical Correlation Analysis (CCA)-based approaches~\cite{chaudhuri2009multi,rupnik2010multi}, whereas recent advances primarily rely on deep architectures~\cite{yan2020higcin,cao2024predictive,bi2024sample,zhan2025elip}. However, most existing methods focus on accuracy rather than predictive reliability, limiting their deployment in safety-critical scenarios. To address this limitation, recent TMVC methods incorporate uncertainty modeling into multi-view classification. Representative examples include Trusted Multi-view Classification (TMC)~\cite{han2020trusted} and Enhanced TMC (ETMC)~\cite{han2022trusted} for complete-view settings, Uncertainty-induced Incomplete Multi-View Data Classification (UIMC)~\cite{xie2023exploring} for incomplete views, and several subsequent variants that improve evidence aggregation or uncertainty modeling~\cite{liu2022trusted,xu2024reliable,liu2024dynamic,yue2025evidential}. However, these methods generally assume that the available views of an instance are semantically consistent. In contrast, our work focuses instead on the more challenging case in which VC may occur during both training and inference.

\subsection{Uncertainty-aware Deep Learning}

Uncertainty modeling is crucial for reliable AI systems and has attracted significant attention~\cite{chen2023uncertainty}. To achieve this, numerous methods are proposed to equip neural networks with uncertainty by adopting distributions over weight parameters, such as Bayesian neural networks~\cite{wang2020survey}, Monte Carlo dropout~\cite{gal2016dropout}, deep ensemble~\cite{lakshminarayanan2017simple}, etc. However, these methods often incur substantial computational cost because they require sampling or multiple forward passes~\cite{lyzhov2020greedy}. To address this, Evidential Deep Learning (EDL) directly infers uncertainty from network outputs via subjective opinions~\cite{gao2025comprehensive}, and has recently been extended to multi-view learning~\cite{xu2024reliable,li2025deep}. Although promising, these methods depend mainly on total evidence and category count, frequently underestimating uncertainty for conflicting instances.
We instead adopt a fuzzy-set-based framework that transforms class-wise memberships into category credibility and subsequently into decision uncertainty via entropy, providing a conflict-aware uncertainty representation tailored for multi-view learning.



\subsection{Learning with View Conflict}

View conflict (VC) represents a special form of cross-view corruption in which different views of the same instance are semantically inconsistent. It was first studied in cross-modal retrieval~\cite{huang2021learning}, followed by a series of methods based on robust losses~\cite{hu2023cross}, or collaborative correction~\cite{wang2025noisy}. Related problems have also been investigated in person re-identification~\cite{qin2024noisy}, multi-view clustering~\cite{sun2025roll}, action recognition~\cite{han2024noise}, video reasoning~\cite{lin2024multi}, and graph matching~\cite{lin2023graph}. Despite this progress, VC has received limited attention in TMVC, where such noisy data may significantly undermine performance. Our work addresses this gap by explicitly modeling inter-view conflict during inference and reducing the adverse impact of conflictive supervision during training.


\begin{figure*}[tbp]
    \centering
    \includegraphics[width=0.9\textwidth]{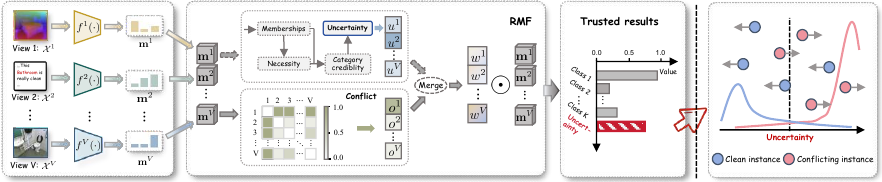}
    \caption{Overview of our R-FUML. First, view-specific networks $\{ f^v(\cdot)\}_{v=1}^{V}$ generate memberships $\{ \mathbf{m}^v \}_{v=1}^{V}$ from each multi-view instance $\{ \mathcal{X}^v\}_{v=1}^{V}$.
    Based on these derived memberships,
    the model further quantifies the intra-view uncertainty $\{u^v \}_{v=1}^V$ and inter-view conflicts $\{o^v \}_{v=1}^V$.
    Next, RMF adaptively computes fusion weights $\{w^v\}_{v=1}^V$ for each view and aggregates the view-specific memberships. Specifically, during training, the fusion weights depend only on conflicts $\{o^v \}_{v=1}^V$, and the estimated uncertainty can be used to distinguish clean from conflicting multi-view instances. During inference, both uncertainty and conflict are jointly incorporated to construct trusted classification results.
    }
    \label{method1}
    \vspace{-0.5cm}
\end{figure*}

\section{Method}


In this section, we first formalize trusted multi-view classification (TMVC) in the presence of view conflict in~\Cref{Problem Definition}. We then elaborate on Fuzzy Set Theory and the corresponding uncertainty quantification mechanism in~\Cref{Uncertainty and Fuzzy Set Theory}, establishing the theoretical foundation for the proposed method. Next, we define the conflict and propose a robust multi-view fusion (RMF) strategy in~\Cref{Conflictive Multi-view Fusion} to mitigate the misleading impact of VC during inference. Finally, to enable model learning under the guidance of VC, we present a dedicated loss function for optimizing category credibility and uncertainty, along with a robust learning against VC (RLVC) framework, in~\Cref{Learning with VC}. 


\subsection{Problem Definition}
\label{Problem Definition}

For clarity, we first introduce the following notations. Let $ \{ \mathcal{X}^v \}_{v=1}^V = \{ \{\mathbf{x}^v_i\}_{i=1}^N \}_{v=1}^V$ denote a $V$-view dataset with $N$ instances, where $\mathbf{x}_i^v$ is the $v$-th view of the $i$-th instance. The corresponding labels are represented as $ \{ \mathcal{Y}^v \}_{v=1}^V = \{ \{\mathbf{y}^v_i\}_{i=1}^N \}_{v=1}^V$. In standard multi-view classification, all views associated with one instance are assumed to be semantically consistent and share the same class label, i.e., $\mathbf{y}^1_i = \mathbf{y}^2_i = ... = \mathbf{y}^V_i$ for the $i$-th instance. In practice, however, this assumption may be violated due to ubiquitous noise~\cite{sun2024robust}, resulting in data corruption or cross-view misalignment. In other words, some views of an instance may be inconsistent with others, i.e., $ y_i^{v_1} \neq y_i^{v_2}, 1\leqslant v_1< v_2\leqslant V$, which is referred to as \emph{view conflict} (VC). Mathematically, the VC problem could be defined as follows:
\begin{definition} $ Let \{ \{\mathbf{x}^v_i\}_{i=1}^N, \{\mathbf{y}^v_i\}_{i=1}^N\}_{v=1}^V$ 
denote a multi-view dataset exhibiting view conflicts. The view conflict (VC) problem is formally defined as follows:
\begin{equation}
    \sum_{v=1}^V \sum_{q\neq v}^V Q ( \mathbf{x}_i^v, \mathbf{x}_i^q ) \leqslant V(V-1), \forall i \in [1, ..., N],
\end{equation}
where $Q$ is an alignment indicator. Specifically, $Q(a, b) = 1$ if two cross-view samples (i.e., $a$ and $b$) are semantically consistent (i.e., not misaligned and with noise), and $Q(a, b) = 0$ otherwise.
\end{definition}

The objective is to learn a trusted multi-view classification model that remains effective when VC is present during both training and inference, while also providing informative uncertainty estimates.





\subsection{Uncertainty and Fuzzy Set Theory}
\label{Uncertainty and Fuzzy Set Theory}

In this subsection, we elaborate on some basic principles and methodology of Fuzzy Set Theory for quantifying the category credibility of each view of a multi-view instance with respect to each category, thereby simultaneously modeling the overall uncertainty of the current prediction.

First proposed by Zadeh in 1965~\cite{zadeh1965fuzzy}, Fuzzy Set Theory relaxes the binary constraints of classical set theory by introducing the concept of graded membership~\cite{liu2010uncertainty}. In other words, it allows individual samples to have partial affiliation with a given set rather than enforcing an absolute `belonging' or `not belonging' dichotomy. This core property endows fuzzy systems with a unique advantage in processing real-world data with ambiguity. Unlike traditional crisp-set-based frameworks, which may struggle with ambiguous or uncertain data distributions, fuzzy systems can naturally characterize and resolve the fuzziness and inherent uncertainty common in practical scenarios. A fundamental metric in Fuzzy Set Theory is the membership degree, hereafter referred to as membership, which is a scalar in $[0, 1]$ quantifying the extent to which a sample is associated with a specific fuzzy set. Notably, this metric exhibits a strong conceptual parallel with the output probabilities of deep classification networks. In standard classification architectures, a network outputs probability values in $[0, 1]$ for each input sample, where each value denotes the likelihood that the sample belongs to a corresponding category, with larger values indicating higher probability. This intrinsic correspondence bridges Fuzzy Set Theory and deep classification networks, allowing us to map the category-wise prediction of a classifier directly to the membership of the sample relative to that category. Based on this, for a given sample $\mathbf{x}^v_i$, its affiliation with all target categories can be represented by a set of memberships. Mathematically, these memberships over all categories are denoted as $m_{i1}^v, m_{i2}^v, \dots, m_{iK}^v$, where $K$ is the number of categories.

However, memberships only characterize the \textbf{possibility measure} that a sample belongs to a specific category, and they fail to capture the \textbf{necessity measure}---the degree of certainty that the sample does not belong to other categories~\cite{liu2010uncertainty}. To compensate for this limitation and enrich the semantic information of category affiliation, we first define the necessity measure for each category based on the membership distribution:
\begin{equation}
    e_{ik}^v = 1 - \max \{m_{il}^v \mid l\neq k\}, \; k=1, ..., K,\\
\end{equation}
where $\max \{m_{il}^v\;|\;l\neq k\}$ denotes the maximum membership of the sample across all categories except the $k$-th category, and $e_{ik}^v$ quantifies the certainty that the sample does not belong to non-$k$ categories. 
Obviously, the larger the value of $e_{ik}^v \in [ 0, 1 ]$, the more certain it is that the sample $\mathbf{x}^v_i$ does not belong to categories other than the $k$-th category.
To integrate the complementary information of possibility and necessity measures, we propose \textbf{category credibility}, which is defined as the arithmetic mean of the possibility and necessity measures. Its definition is given below:
\begin{definition}
Let $\mathbf{m}_i^v = [m_{i1}^v,m_{i2}^v,...,m_{iK}^v]$ be the membership vector of the $i$-th sample in the $v$-th view, where $\forall m_{ik}^v \in[0,1],\; k=1,2,..., K$. The category credibility of the $i$-th sample with respect to the $k$-th category is defined as
\begin{equation}
    c_{ik}^v = \frac{1}{2}(m_{ik}^v + 1 - \max \{m_{il}^v\mid l\neq k\}),\; k=1,2,...,K.\\
\label{Eq:cr}
\end{equation}
The category credibility can be organized as a vector $\mathbf{c}_i^v = [c_{i1}^v, c_{i2}^v, ..., c_{iK}^v] \in \mathbb{R}^{K}$.
\label{def:category credibility}
\end{definition}

Because $m_{ik}^v \in [0,1]$ and $e_{ik}^v \in [0,1]$, we have $c_{ik}^v \in [0,1]$.
Although the category credibility effectively reflects the uncertainty of a single category prediction, it does not qualify the uncertainty of the entire decision results. To address this issue, we introduce a decision uncertainty metric inspired by Shannon entropy, which can capture uncertainty arising from information deficiency. The uncertainty is defined as follows:
\begin{definition}
Let $\mathbf{c}_i^v = [c_{i1}^v,c_{i2}^v,...,c_{iK}^v]$ be the category credibility vector of the $i$-th sample in the $v$-th view, where $\forall c_{ik}^v \in[0,1],\; k=1,2,..., K$. The uncertainty is defined as
\begin{equation}
\begin{aligned}
    u_i^v &= \frac{\sum^K_{k=1} H(c^v_{ik})}{K\cdot \ln 2}\\
    &= \frac{\sum^K_{k=1} -c_{ik}^v\cdot ln(c_{ik}^v) - (1-c_{ik}^v)\cdot ln(1-c_{ik}^v)}{K\cdot \ln 2},
\end{aligned}
\label{Eq:Uncer}
\end{equation}
where $K$ is the number of categories and $H(c^v_{ik})$ is the entropy of category credibility $c_{ik}^v$. This uncertainty ranges from $0$ to $1$, with larger values indicating greater uncertainty in the sample’s prediction result.
\label{def:uncertainty}
\end{definition}

\textbf{Difference from Existing Classifiers.} 
\textbf{First}, traditional DNN-based classifiers use a softmax output layer, which often leads to overconfidence in predictions~\cite{gao2025comprehensive}. In contrast, we model the outputs of classification networks as fuzzy memberships, which can be further derived into category credibility and decision uncertainty, thereby mitigating the over-confidence issue. To intuitively illustrate this, consider a three-class classification task with the membership degree vector $\mathbf{m} = [0.80,0.10,0.10]$. The uncertainty calculated directly from these memberships is $0.55$. In our framework, we first compute the category credibility vector as $\mathbf{c} = [0.85,0.15,0.15]$, which yields a higher uncertainty value of $0.61$ and provides a more conservative and realistic uncertainty estimation—effectively mitigating the overconfidence. \textbf{Second}, in EDL-based classifiers~\cite{sensoy2018evidential}, the uncertainty estimation mechanism relies entirely on total evidence and the number of categories, ignoring the per-category evidence distribution. This may lead to inaccurate uncertainty quantification~\cite{duan2025fuzzy}. In our model, decision uncertainty is measured by the normalized entropy of category credibility, which explicitly accounts for the distribution of memberships across all categories and enables more precise uncertainty characterization.

\subsection{Conflictive Multi-view Fusion}
\label{Conflictive Multi-view Fusion}

After quantifying uncertainty for each view, we address multi-view/multi-modal data adaptation—a scenario inherently plagued by inter-view conflicts. Such conflicts primarily stem from environmental perturbations (e.g., sensor malfunctions, adverse weather, or data communication anomalies), which introduce noisy or misaligned views into multi-view data~\cite{zhang2024multimodal}. Resolving these conflicts is critical for improving the accuracy and robustness of MVC because the two types of conflicting views pose distinct challenges: i) noisy views typically exhibit high uncertainty, which undermines accurate decision-making and may cause erroneous outcomes; ii) misaligned views often produce highly conflicting but low-uncertainty predictions, which may mislead the final classification decision. Therefore, during multi-view fusion, the impact of both noisy and misaligned views should be reduced so that cleaner, aligned, and more reliable views contribute more strongly. We first define conflict between views and then use the defined uncertainty and conflict to construct a multi-view fusion strategy that is robust to VC.



\subsubsection{\textbf{Conflict Inference}} The uncertainty defined above does not allow for the assessment of inconsistencies between different views. To address this, we define inter-view conflict as follows:

\begin{definition}
Let $\{\mathbf{m}^v_i\}_{v=1}^{V}$ denote the set of multi-view membership vectors for the $i$-th multi-view instance. The conflict of the $v$-th view relative to other views is defined as:
\begin{equation}
    o^v_i = \frac{1}{V-1} \sum_{j\neq v}^{V} \left( 1 - \frac{\mathbf{m}^v_i \cdot \mathbf{m}_i^j}{ ||\mathbf{m}_i^v|| \cdot ||\mathbf{m}_i^j||} \right).
\label{credibility degree uncertainty}
\end{equation}
For nonzero membership vector $\mathbf{m}^v_i$ with all elements in $[0,1]$, the conflict $o^v_i$ lies in $[0,1]$. The larger the value, the more conflicting the view $v$ is with the other views.
\label{def:conflictive degree}
\label{def:conflictive degree}
\end{definition}

Views with relatively high conflict are considered less aligned with the other views, whereas views with relatively low conflict should contribute more to fusion. Next, we leverage view-specific uncertainty and inter-view conflict to propose a robust multi-view fusion strategy that enables reliable fusion in the presence of noisy and misaligned views.

\subsubsection{\textbf{Robust Multi-view Fusion (RMF)}} In multi-view decision-level fusion, we aim to fuse views with low uncertainty and low conflict. Following \Cref{def:uncertainty} and \Cref{def:conflictive degree}, the final memberships $\mathbf{m}^a_i$ from different views are fused as follows:
\begin{equation}
    w^v_i =  \frac{g\left((1-u^v_i)(1-o^v_i)\right)}{\sum_{v=1}^V g\left((1-u^v_i)(1-o^v_i)\right)},\quad
    \mathbf{m}^a_i = \sum_{v=1}^V w^v_i \cdot \mathbf{m}^v_i,
\label{Eq:RMF}
\end{equation}
where $g(\cdot)$ is a monotonically increasing function. In this work, we use the exponential function $exp(\cdot)$ as $g(\cdot)$. With RMF, we can obtain the final membership of each category and then infer the overall uncertainty using \Cref{Eq:Uncer}. 
In addition, it's worth mentioning that RMF is effective only when both uncertainty and conflict are accurately estimated. During training, RMF may negatively affect the model optimization when VC is present. 

The reason is as follows. Without loss of generality, we assume that the view $\mathcal{X}^A$ is misaligned with the other views. Because of incorrect supervised information, the category credibility during training of view $\mathcal{X}^A$ may be incorrectly calculated during training. Consequently, the uncertainty of view $\mathcal{X}^A$, i.e., $u^A$, may also be incorrectly estimated, resulting in inaccurate multi-view fusion. In contrast, conflict $o^A$ is computed only from neural network outputs, i.e., memberships ($\mathbf{m}^A$), and does not directly use any supervised information; therefore, it is less directly affected by VC. Accordingly, during training, we fuse memberships through 
\begin{equation}
    \hat{w}^v_i = \frac{g\left(1-o^v_i\right)}{ \sum_{v=1}^V g\left(1-o^v_i\right)},\quad \mathbf{m}^a_i = \sum_{v=1}^V \hat{w}^v_i \cdot \mathbf{m}^v_i,
\end{equation}
whereas \Cref{Eq:RMF} is used during testing.


\textbf{Intuitive Explanation of the Effectiveness of RMF.} Without loss of generality, we assume that view $\mathcal{X}^A$ is clean, view $\mathcal{X}^B$ is noisy due to unknown environmental factors or sensor failure, and view $\mathcal{X}^C$ is misaligned with the other views for similar reasons. In this case, $\mathcal{X}^A$ aligns with the distribution of clean training data, whereas $\mathcal{X}^B$ deviates significantly from it. Accordingly, we have $u^A \leqslant u^B$ and $o^A \leqslant o^B$, leading to $w^A \geqslant w^B$. Therefore, in our R-FUML framework, the multi-view decision tends to rely more on the high-quality view $\mathcal{X}^A$ than on the noisy view $\mathcal{X}^B$. In addition, although the uncertainty of view $\mathcal{X}^C$ is not as high as that of view $\mathcal{X}^B$, its relative view $\mathcal{X}^C$ has higher conflict with the other views. Accordingly, we have $u^A \approx u^C$ and $o^A \leqslant o^C$, and thus $w^A \geqslant w^C$. Therefore, the multi-view decision tends to rely more on the low-conflict view $\mathcal{X}^A$ than on $\mathcal{X}^C$. As for the weight between views $\mathcal{X}^B$ and view $\mathcal{X}^C$, it is related to the careful consideration of its uncertainty and conflict. By dynamically determining the fusion weights of each view, R-FUML effectively mitigates the influence of noisy and misaligned views, i.e., conflicting views, thereby improving robust classification for multi-view instances with VC.


\subsection{Learning with VC}
\label{Learning with VC}

After constructing the model framework, we now focus on model training under VC. To achieve this, we first design a loss function tailored for category credibility optimization. Subsequently, we propose a robust training paradigm against VC, consisting of four sequential stages.



\subsubsection{\textbf{Category Credibility Learning}}
To learn category credibility and uncertainty, we first model the outputs of a neural network as memberships. Specifically, we adopt a two-step transformation process. First, we apply $L_p -$normalization to the logits to constrain their values to $[-1, 1]$. Second, we employ the ReLU activation function to map these normalized values to $[0, 1]$. The resulting values are then treated as the memberships corresponding to each category:
\begin{equation}
    \mathbf{m}^v_i = \text{ReLU}(\frac{\mathbf{a}^v_i}{||\mathbf{a}^v_i||_p}),
\label{Eq: fuzzy layer}
\end{equation}
where $\mathbf{a}^v_i \in \mathbb{R}^{K}$ denotes the logits of the neural network. The corresponding category credibility vector $\mathbf{c}_i^v \in \mathbb{R}^{K}$ can then be derived by \Cref{Eq:cr}.

For discriminative learning, each sample should have the maximal category credibility for its matched category and minimal category credibility for all unmatched categories. Intuitively, this objective can be achieved by aligning the category credibility $\mathbf{c}^v_i$ with the corresponding one-hot label $\mathbf{y}^v_i$, for example, by minimizing the mean absolute error ($|\mathbf{c}^v_i - \mathbf{y}^v_i|$), 
mean squared error ($\|\mathbf{c}^v_i - \mathbf{y}^v_i\|_2$), or 
cross-entropy loss ($-\mathbf{y}^v_i \cdot \log(\mathbf{c}^v_i) - (1-\mathbf{y}^v_i) \cdot \log(1-\mathbf{c}^v_i)$). However, these strategies risk over-optimizing the necessity $\mathbf{e}^v_i$ of unmatched categories, causing the neural network to converge to a local optimum. The underlying reason is as follows. When label $y_{ik}^v = 0$, the memberships $m_{ik}^v$ tends toward 0; consequently, the necessity $e_{ik}^v = 1 - \max \{m_{il}^v \mid l \neq k \}$ also approaches 0, which incorrectly drives $m_{il}^v$ toward 1. This is problematic because $m_{il}^v$ should approach 0 (rather than 1) when $y_{il}^v = 0$. To address this issue, we propose a tailored category credibility learning loss to optimize the category credibility, thereby guiding the model toward the correct optimization direction:
\begin{equation}
    \mathcal{L}_{ccl} = \frac{1}{N_b} \sum_{i=1}^{N_b}  -\mathbf{y}_i^v \cdot \log (\mathbf{r}_i^v) - (1-\mathbf{y}_i^v)  \cdot \log(1-\mathbf{r}_i^v),
\label{ccl}
\end{equation}
where $N_b$ is the batch size,  and $\mathbf{r}_i^v = \phi^{tr}(\mathbf{m}_i^v,\mathbf{y}_i^v)=[r_{i1}^v, r_{i2}^v, ..., r_{iK}^v] \in \mathbb{R}^{K}$ represents the category credibility during training. The component $r_{ik}^v$ is formulated as
\begin{equation}
\small
\begin{aligned}
    r_{ik}^v= 
    \begin{cases}
        \dfrac{m_{ik}^v + 1 - \max \{m_{il}^v \mid l \neq k \}}{2}, \;\;\;\;\;\;\;\;\;\;\;\;\;\;\;\;\;
             \text{if } y_{ik}^v = 1,& \\[2ex]  
        \dfrac{m_{ik}^v + 1 - m_{il}^v}{2}, \;\;\;\;\;\;\;\;\;\;\;
             \text{if } y_{ik}^v = 0, \ l = \underset{k}{\arg \max}\;\;y_{ik}^v,&
    \end{cases}
\end{aligned}
\label{Eq:credibility degrees during training}
\end{equation}
where $k = 1, 2, ..., K$. As indicated in \Cref{ccl}, by leveraging label information, this loss function ensures that $m_{ik}^v$ approaches 1 for matched categories ($y_{ik}^v=1$) and 0 for unmatched categories ($y_{ik}^v=0$). Specifically, when $y_{ik}^v = 0$ and $y_{il}^v = 1$, where $l = \underset{k}{\arg \max} \; y_{ik}^v$, the loss enforces the memberships of the matched category to be larger than that of any unmatched category after training. For matched categories, the training-phase necessity should be calculated as $1 - \max \{m_{il}^v \mid l \neq k \}$, which drives $\max \{m_{il}^v \mid l \neq k \}$ toward 0. For unmatched categories, the training-phase necessity should be computed as $1 - m_{il}^v$, forcing $m_{il}^v$ to approach 1 rather than 0. 
Therefore, this approach effectively regulates the optimization of necessity measures and category credibility, avoiding over-optimization and supporting correct model convergence.

\begin{figure*}[htbp]
    \vspace{-0.1cm}
    \centering
    \begin{subfigure}[b]{0.22\textwidth}
        \centering
        \includegraphics[width=\textwidth]{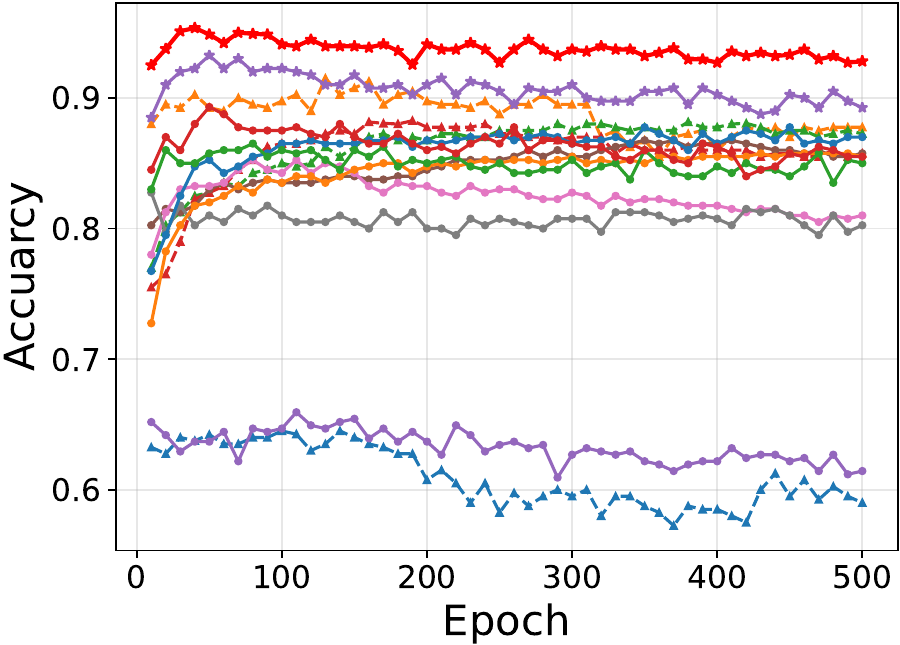}
        \caption{HW}
    \end{subfigure}
    \hfill
    \begin{subfigure}[b]{0.22\textwidth}
        \centering
        \includegraphics[width=\textwidth]{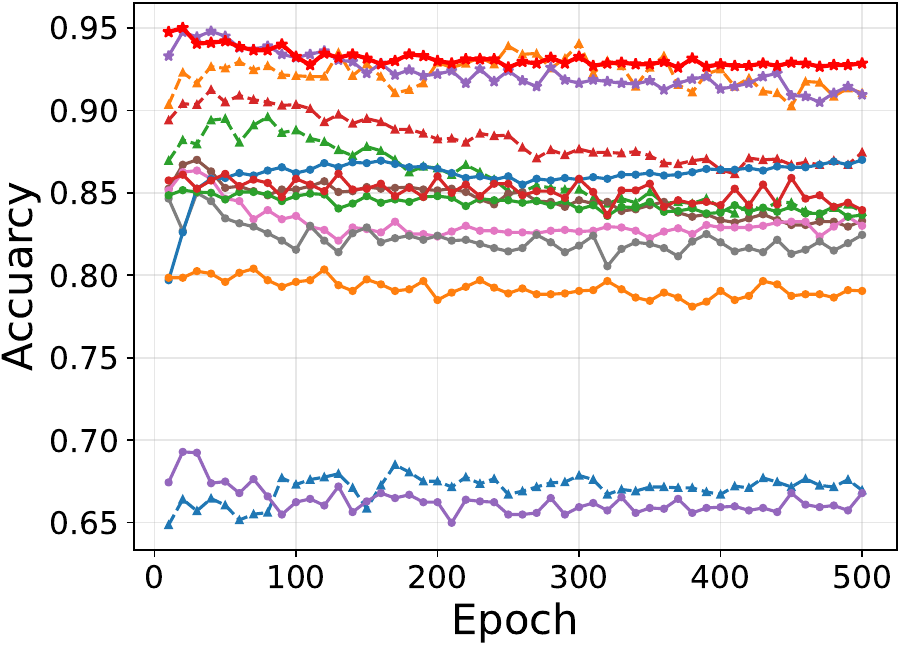}
        \caption{Fashion}
    \end{subfigure}
    \hfill
    \begin{subfigure}[b]{0.22\textwidth}
        \centering
        \includegraphics[width=\textwidth]{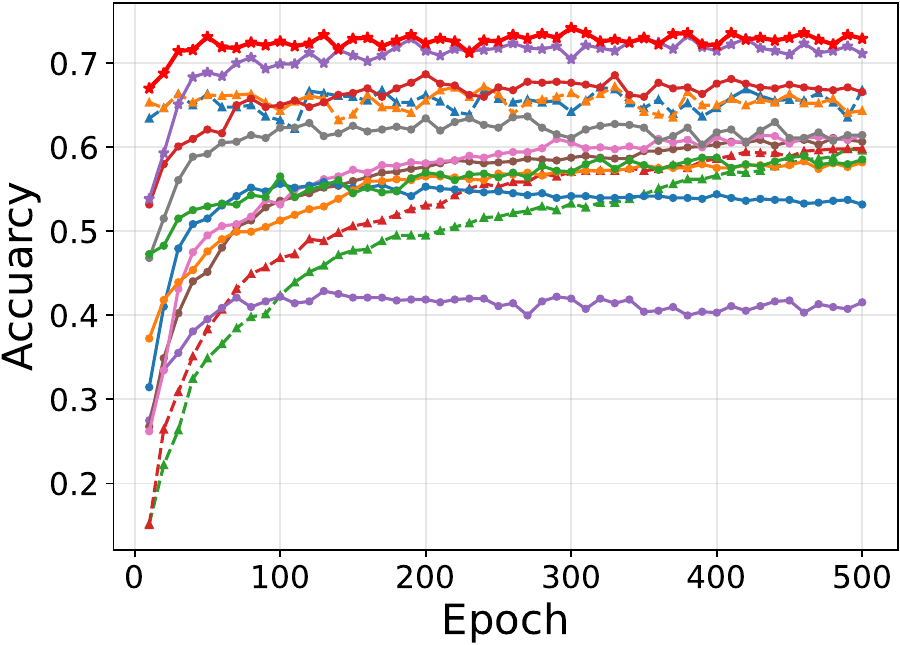}
        \caption{Scene}
    \end{subfigure}
    \hfill
    \begin{subfigure}[b]{0.27\textwidth}
        \centering
        \includegraphics[width=\textwidth]{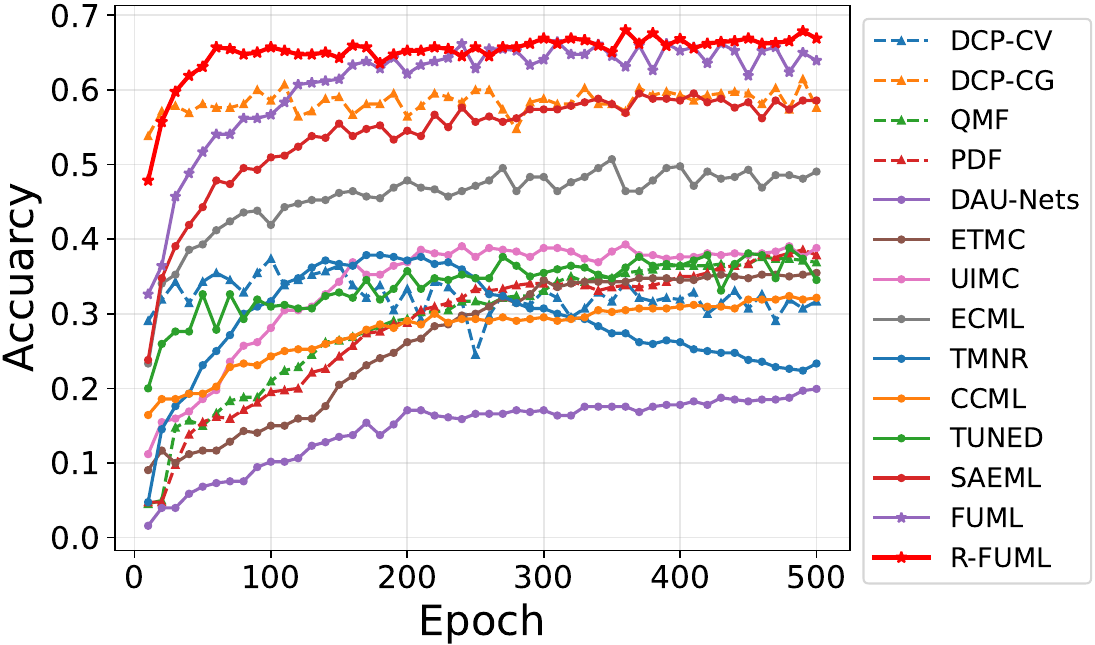}
        \caption{LandUse}
    \end{subfigure}
    \vspace{-0.1cm}
    \caption{Test accuracy versus epochs on HW, Fashion, Scene, and LandUse datasets with $40\%$ VC in both training and test sets.}
    \label{fig:robustness study}
    \vspace{-0.6cm}
\end{figure*}




%

\subsubsection{\textbf{Robust Learning Against VC (RLVC)}} To alleviate the negative impact of VC, a straightforward way is to filter conflicting views in training samples. This can directly reduce incorrect supervision information and thereby enhance robustness. Prior studies on learning with noisy labels~\cite{li2020dividemix,yang2024robust} depend on the \textit{memorization effect}~\cite{tan2021co} of DNNs for noise filtering. This effect refers to the phenomenon that noisy or hard samples tend to have higher losses than clean or easy samples in early training. However, the discriminative power of this phenomenon degrades as training progresses, since DNNs eventually memorize noisy samples, causing their losses to drop and overlap with those of clean samples~\cite{kim2024learning}. To sustain reliable sample identification, we adopt a cyclical training paradigm inspired by \cite{huang2019o2u}, which causes the network to transition from underfitting to overfitting multiple times during training, thereby effectively identifying conflicting samples. The stages are as follows:

\textbf{Stage 1: Pre-training.} We first train the network directly on the original dataset (including VC). The goal is to enable all views to simultaneously form accurate decisions and thereby improve overall performance. To this end, we employ a multi-task strategy with the following overall loss function:

\begin{equation}
    \mathcal{L}_{total} =  \gamma \cdot \mathcal{L}_{ccl}(\mathbf{r}^{a}, \mathbf{y}) + \sum_{v=1}^V \mathcal{L}_{ccl}(\mathbf{r}^{v}, \mathbf{y}).
\label{total loss function}
\end{equation}
In early training, conflict cannot be accurately estimated, which may cause RMF to fail.
Consequently, $\mathcal{L}_{ccl}(\mathbf{r}^{a}, \mathbf{y})$ should be assigned a low initial weight that gradually increases as training progresses.
To achieve this, we introduce a dynamic coefficient $\gamma = \min(t/N_w, 1)$, where $t$ is the current epoch, and $N_w$ is the number of warm-up epochs.
Meanwhile, to mitigate the noise introduced by VC in partial views, we employ a large batch size ($b_l$) at this stage, following~\cite{rolnick2017deep}. We continue training until overfitting is observed.

\textbf{Stage 2: Cyclical Training.}
To accumulate sufficient loss statistics for subsequent conflicting-view identification and division, we resume training with a smaller batch size ($b_s$) and a cyclical learning rate for multiple cycles. Specifically, let $\eta_{max}$ and $\eta_{min}$ denote the maximum and minimum learning rates, respectively, with $\eta_{max} > \eta_{min}$. 
The learning rate at the $t$-th epoch, denoted by $\eta_t$, follows a linear decay schedule:
\begin{equation}
\begin{aligned}
    &\eta_t = s(t)\times \eta_{min} + (1-s(t))\times \eta_{max}, \\
    &s(t) = \frac{(t-1) \mod N_c}{N_c-1},
\end{aligned}
\label{eq:Cyclical Training}
\end{equation}
where $N_c$ is the number of epochs per cycle. Model optimization also follows~\Cref{total loss function}. After this stage, we compute the per-view average loss for each multi-view instance across all epochs.

\textbf{Stage 3: Conflictive View Division.} After collecting sufficient loss statistics, we identify and partition conflicting views. Following~\cite{qin2024noisy}, we exploit the Expectation-Maximization (EM) algorithm to fit a two-component Gaussian Mixture Model (GMM) to the per-view average losses and compute the posterior probability $p(k|\overline{\mathcal{L}^v_i}) = \frac{p(k)p(\overline{\mathcal{L}^v_{i}}|k)}{p(\overline{\mathcal{L}^v_{i}})}$ for the $v$-th view of the $i$-th multi-view instance. This posterior probability serves as the score for classifying the sample as clean or conflicting, where $k \in \{0,1\}$  indicates clean versus conflicting. The average loss of the $v$-th view of the $i$-th multi-view instance across all cyclical training epochs, denoted by $\overline{\mathcal{L}^v_{i}}$, is computed as:
\begin{equation}
\overline{\mathcal{L}^v_{i}} = \frac{1}{N_{e2}} \sum_{t=1}^{N_{e2}} (-\mathbf{y}_i^v \cdot \log (\mathbf{r}_i^{vt}) - (1-\mathbf{y}_i^v)  \cdot \log(1-\mathbf{r}_i^{vt})),
\end{equation}
where $\mathbf{r}_i^{vt}$ denotes the training-phase category credibility of of the $v$-th view of the $i$-th multi-view instance at the $t$-th epoch, and $N_{e2}$ is the maximal number of epochs in Stage 2.
To obtain a binary partition, we then apply a threshold $\beta = 0.5$ to the posterior probability set $\{p(k=0|\overline{\mathcal{L}^v_{i}})\}_{i=1}^N$, yielding clean and conflicting subsets for each view:
\begin{equation}
\begin{aligned}
    \mathcal{X}^v_{\text{cle}} = \{ \mathbf{x}_i^v | p(k=0|\overline{\mathcal{L}^v_i})>\beta , \forall \mathbf{x}_i^v \in \mathcal{X}^v\}, \\ 
    \mathcal{X}^v_{\text{con}} = \{ \mathbf{x}_i^v | p(k=0|\overline{\mathcal{L}^v_i})\le \beta , \forall \mathbf{x}_i^v \in \mathcal{X}^v\}, 
\end{aligned}
\label{eq: clean noisy division}
\end{equation}
where $\mathcal{X}^v_{\text{cle}}$ and $\mathcal{X}^v_{\text{con}}$ are the divided clean and conflicting sets of the $v$-th view, respectively. Rather than discarding conflicting samples entirely, we preserve them with adaptive importance weights that reflect their estimated reliability:
\begin{equation}
\begin{aligned}
    d_{i}^v = 
    \begin{cases}
        1, & \text{if } \mathbf{x}^v_{i} \in \mathcal{X}^v_{\text{cle}}, \\
        p^v_i, & \text{if } \mathbf{x}^v_{i} \in \mathcal{X}^v_{\text{con}}.
    \end{cases}
\end{aligned}
\label{eq: sample importance}
\end{equation}
Here, a larger $d_i^v$ indicates greater importance, ensuring that potentially useful supervisory signals from borderline samples are not prematurely discarded. $p^v_i$ is short for $p(k=0|\overline{\mathcal{L}^v_i})$.


\textbf{Stage 4: Robust Learning.} Finally, to mitigate the impact of VC, we re-initialize the network parameters and re-train the model until the loss stabilizes by penalizing conflicting views. To this end, we rewrite the total loss function as:
\begin{equation}
\begin{aligned}
    &\mathcal{L}_{total\_rl} = \gamma \cdot \mathcal{L}_{ccl}(\mathbf{r}^{a}, \mathbf{y}) + \sum_{v=1}^V \mathcal{L}_{rccl}(\mathbf{r}^{v}, \mathbf{y}, \mathbf{d}^{v}),\\
    & \mathcal{L}_{rccl} = \frac{1}{N_b} \sum_{i=1}^{N_b} d_i^v \cdot (- \mathbf{y}_i^v \cdot \log (\mathbf{r}_i^v) - (1-\mathbf{y}_i^v)  \cdot \log(1-\mathbf{r}_i^v)),
\end{aligned}
\label{eq: robust total function}
\end{equation}
where $\mathbf{d}^{v} = \{d^v_i \}_{i=1}^N$.

\begin{table*}[htbp]
\setlength\tabcolsep{0.5pt}
\caption{
Accuracy (\%) under varying VC rates in the training and test sets. 
Best results are in \textbf{bold}, and second-best results are \underline{underlined}. 
Statistics are reported as mean$\pm$std over 10 runs. 
The symbol `$\bullet$' indicates trusted methods that provide decision uncertainty.
}
\vspace{-0.3cm}
\tabcolsep=0.15cm 
\begin{center}
\resizebox{1.\textwidth}{!}{%
\begin{tabular}{p{0.52cm}|p{1.5cm}|c|cccccccc}
\toprule
VC&Methods&Ref.&HW & MSRC & NUSOBJ &Fashion&Scene& LandUse & Leaves &PIE \\
\midrule
\multirow{14}{*}{0\%}
&DCP-CV&TPAMI'22&98.75$\pm$0.59& 92.86$\pm$2.61&32.19$\pm$9.48&97.96$\pm$0.16&76.70$\pm$2.15&71.71$\pm$2.09&95.62$\pm$1.38&86.32$\pm$4.87\\
&DCP-CG&TPAMI'22&99.00$\pm$0.47&95.24$\pm$3.69&43.65$\pm$1.10& 98.11$\pm$0.23&77.79$\pm$1.73&75.74$\pm$0.98&98.19$\pm$0.46&90.59$\pm$1.99\\
&QMF&ICML'23&98.72$\pm$0.48&97.86$\pm$1.28&45.41$\pm$0.43&98.93$\pm$0.32&68.58$\pm$1.49&47.86$\pm$2.55&95.69$\pm$1.25&92.06$\pm$1.64\\
&PDF&ICML'24&98.40$\pm$0.37&97.14$\pm$1.78&46.78$\pm$0.33&98.95$\pm$0.19&70.25$\pm$1.21&45.17$\pm$2.66&98.03$\pm$0.71&92.57$\pm$1.66\\
&DUA-Nets$\bullet$&AAAI'21&98.10$\pm$0.32&84.67$\pm$3.03&27.75$\pm$0.00&91.08$\pm$0.17&65.01$\pm$1.55&45.24$\pm$1.85&90.31$\pm$1.25&90.56$\pm$0.47\\
&ETMC$\bullet$&TPAMI'22&98.75$\pm$0.00& 92.86$\pm$3.01&44.23$\pm$0.76&96.21$\pm$0.36&71.61$\pm$0.28& 43.52$\pm$3.19&91.44$\pm$2.39&93.75$\pm$1.08 \\
&UIMC$\bullet$&CVPR'23&98.25$\pm$0.00&98.81$\pm$1.19&43.42$\pm$0.12&98.13$\pm$0.13&77.70$\pm$0.00&57.95$\pm$0.61&95.31$\pm$0.71&91.69$\pm$2.16 \\
&ECML$\bullet$&AAAI'24&98.72$\pm$0.39& 94.05$\pm$1.60&42.62$\pm$0.42&97.93$\pm$0.35&76.19$\pm$0.12&60.10$\pm$2.01&92.53$\pm$1.94&94.71$\pm$0.02\\
&TMNR$\bullet$&IJCAI'24&97.20$\pm$0.63&94.05$\pm$3.24&34.52$\pm$0.85&94.10$\pm$0.50&68.10$\pm$1.15&27.38$\pm$1.88&90.13$\pm$1.53&89.53$\pm$1.89\\
&CCML$\bullet$&MM'24&97.60$\pm$0.62&96.90$\pm$2.39&41.43$\pm$0.71&95.16$\pm$0.41&73.87$\pm$1.83&60.86$\pm$1.93&97.72$\pm$0.92&93.97$\pm$1.67\\
&TUNED$\bullet$&AAAI'25&98.28$\pm$0.78&97.14$\pm$2.08&37.61$\pm$0.26&96.06$\pm$0.37&74.18$\pm$1.51&49.81$\pm$2.00&93.59$\pm$1.08&94.49$\pm$2.57\\
&SAEML$\bullet$&MM'25& 98.87$\pm$0.46&98.81$\pm$1.60&43.22$\pm$0.48&98.26$\pm$0.31&81.22$\pm$0.82&75.07$\pm$1.47&97.94$\pm$0.92&93.97$\pm$1.67\\
&FUML$\bullet$&ICML'25&\textbf{99.20$\pm$0.36}&\underline{99.76$\pm$0.75}&\textbf{48.23$\pm$0.42}&\underline{98.96$\pm$0.25}&\underline{79.41$\pm$1.34}&\underline{76.71$\pm$0.46}&\textbf{99.78$\pm$ 0.27}&\textbf{96.18$\pm$1.24}\\
&R-FUML$\bullet$&Ours&\underline{99.12$\pm$0.32}&\textbf{99.83$\pm$1.20}&\underline{46.89$\pm$0.41}&\textbf{99.02$\pm$0.33}&\textbf{79.46$\pm$0.79}&\textbf{77.00$\pm$1.78}&\underline{99.69$\pm$0.24}&\underline{95.65$\pm$1.37}\\
\midrule
\multirow{14}{*}{20\%}&DCP-CV&TPAMI'22&94.38$\pm$0.58&76.67$\pm$3.33&38.52$\pm$0.78&95.16$\pm$0.28&71.28$\pm$0.63&66.95$\pm$0.85&45.28$\pm$2.11&62.79$\pm$1.68\\
&DCP-CG&TPAMI'22&93.95$\pm$0.51&83.33$\pm$2.13&38.79$\pm$0.98&95.49$\pm$0.26&69.68$\pm$0.83&62.88$\pm$1.05&52.69$\pm$3.32&63.46$\pm$1.74\\
&QMF&ICML'23&93.15$\pm$1.30&90.71$\pm$3.60&42.92$\pm$0.63&93.44$\pm$0.36&63.98$\pm$0.93&42.19$\pm$2.31&80.47$\pm$1.70&80.44$\pm$1.55\\
&PDF&ICML'24&93.15$\pm$0.84&91.90$\pm$6.49&42.81$\pm$0.55&94.98$\pm$0.42&64.29$\pm$0.88&42.71$\pm$2.26&83.50$\pm$1.64&79.63$\pm$3.22\\
&DUA-Nets$\bullet$&AAAI'21&89.15$\pm$0.78&76.90$\pm$5.74&25.82$\pm$0.37&85.78$\pm$0.52&48.62$\pm$0.98&40.10$\pm$1.51&74.16$\pm$1.57&57.06$\pm$3.40\\
&ETMC$\bullet$&TPAMI'22&92.82$\pm$1.01&88.33$\pm$2.25&39.77$\pm$0.58&91.76$\pm$0.49&65.44$\pm$1.32&39.64$\pm$2.07&82.34$\pm$1.41&79.63$\pm$3.22\\
&UIMC$\bullet$&CVPR'23&92.28$\pm$1.13&92.38$\pm$3.81&38.24$\pm$0.58&91.48$\pm$0.21&66.80$\pm$1.16&43.10$\pm$2.23&81.00$\pm$1.35&78.75$\pm$2.84\\
&ECML$\bullet$&AAAI'24&90.95$\pm$0.93&83.10$\pm$3.10&38.87$\pm$0.38&91.11$\pm$0.45&69.01$\pm$0.97&54.12$\pm$2.06&84.09$\pm$1.51&78.16$\pm$2.65\\
&TMNR$\bullet$&IJCAI'24&91.90$\pm$1.08&92.14$\pm$4.40&22.86$\pm$0.40&90.16$\pm$0.54&58.21$\pm$1.02&42.62$\pm$1.97&70.97$\pm$2.84&82.21$\pm$2.39\\
&CCML$\bullet$&MM'24&91.80$\pm$1.16&91.43$\pm$4.54&35.29$\pm$0.49&90.46$\pm$0.46&63.68$\pm$0.99&36.26$\pm$1.40&82.28$\pm$1.82&82.43$\pm$2.72\\
&TUNED$\bullet$&AAAI'25&91.78$\pm$0.98&88.57$\pm$2.56&11.94$\pm$0.33&90.13$\pm$0.47&63.00$\pm$2.01&43.21$\pm$1.70&80.19$\pm$1.54&81.99$\pm$3.29\\
&SAEML$\bullet$&MM'25&94.42$\pm$1.22&94.76$\pm$4.10&38.94$\pm$0.51&93.19$\pm$0.47&73.13$\pm$0.81&65.60$\pm$1.51&86.41$\pm$1.52&82.87$\pm$2.21\\
&FUML$\bullet$&ICML'25&\underline{96.62$\pm$0.82}&\underline{95.95$\pm$1.52}&\textbf{43.55$\pm$0.44}&\underline{96.72$\pm$0.45}&\underline{75.66$\pm$1.09}&\underline{70.43$\pm$1.80}&\underline{96.03$\pm$1.10}&\underline{86.91$\pm$2.07}\\
&R-FUML$\bullet$&Ours&\textbf{97.05$\pm$1.04}&\textbf{96.90$\pm$3.99}&\underline{43.44$\pm$0.43}&\textbf{96.84$\pm$0.30}&\textbf{76.58$\pm$0.91}&\textbf{70.74$\pm$1.87}&\textbf{96.09$\pm$0.92}&\textbf{87.13$\pm$2.33}\\
\midrule
\multirow{14}{*}{40\%}&DCP-CV&TPAMI'22&89.47$\pm$0.39&66.19$\pm$3.50&35.64$\pm$0.96&93.22$\pm$0.36&66.42$\pm$0.37&59.74$\pm$1.03&37.38$\pm$1.66&56.32$\pm$1.10\\
&DCP-CG&TPAMI'22&91.88$\pm$0.52&75.00$\pm$3.41&35.14$\pm$1.11&94.06$\pm$0.38&65.16$\pm$0.56&60.31$\pm$0.67&43.22$\pm$2.68&58.09$\pm$1.64\\
&QMF&ICML'23&88.15$\pm$1.30&82.62$\pm$3.85&39.88$\pm$0.42&89.53$\pm$0.71&59.80$\pm$1.18&37.90$\pm$2.33&72.12$\pm$2.79&69.78$\pm$2.07\\
&PDF&ICML'24&88.12$\pm$1.05&84.05$\pm$6.65&39.65$\pm$0.32&91.35$\pm$0.37&59.86$\pm$1.16&38.33$\pm$1.94&73.00$\pm$2.14&67.79$\pm$1.76\\
&DUA-Nets$\bullet$&AAAI'21&81.63$\pm$2.03&71.43$\pm$4.52&22.82$\pm$0.21&80.59$\pm$0.60&43.58$\pm$1.17&35.55$\pm$2.56&62.19$\pm$2.16&43.01$\pm$2.59\\
&ETMC$\bullet$&TPAMI'22&87.02$\pm$1.59&80.95$\pm$4.64&37.06$\pm$0.43&87.02$\pm$0.38&61.07$\pm$1.21&36.48$\pm$2.31&73.72$\pm$1.65&69.49$\pm$2.95\\
&UIMC$\bullet$&CVPR'23&85.20$\pm$0.86&83.57$\pm$5.27&35.91$\pm$0.36&86.37$\pm$0.63&62.64$\pm$1.03&39.93$\pm$2.57&73.81$\pm$2.53&68.82$\pm$2.93\\
&ECML$\bullet$&AAAI'24&84.72$\pm$1.83&72.14$\pm$6.74&36.10$\pm$0.36&85.94$\pm$0.62&64.67$\pm$1.15&50.24$\pm$1.95&77.16$\pm$1.74&67.13$\pm$3.17\\
&TMNR$\bullet$&IJCAI'24&87.03$\pm$1.28&85.71$\pm$4.99&21.62$\pm$0.64&86.79$\pm$0.79&55.27$\pm$1.12&39.69$\pm$1.70&60.41$\pm$2.39&73.53$\pm$3.01\\
&CCML$\bullet$&MM'24&86.35$\pm$1.39&84.05$\pm$5.22&32.54$\pm$0.42&86.00$\pm$0.58&58.97$\pm$1.05&33.45$\pm$2.20&73.63$\pm$1.54&72.28$\pm$3.31\\
&TUNED$\bullet$&AAAI'25&86.25$\pm$1.32&79.29$\pm$3.20&11.95$\pm$0.34&85.16$\pm$0.65&59.31$\pm$1.66&39.69$\pm$1.70&73.00$\pm$1.75&72.72$\pm$2.99\\
&SAEML$\bullet$&MM'25&89.30$\pm$1.27&85.95$\pm$4.70&36.52$\pm$0.44&88.16$\pm$0.69&68.75$\pm$0.92&60.12$\pm$2.22&79.56$\pm$1.80&72.13$\pm$4.44\\
&FUML$\bullet$&ICML'25&\underline{93.35$\pm$1.51}&\underline{88.57$\pm$4.93}&\underline{40.49$\pm$0.41}&\underline{94.82$\pm$0.42}&\underline{73.27$\pm$0.79}&\underline{66.12$\pm$2.04}&\underline{92.53$\pm$1.14}&\underline{76.18$\pm$1.81}\\
&R-FUML$\bullet$&Ours&\textbf{95.35$\pm$1.22}&\textbf{89.52$\pm$3.87}&\textbf{40.65$\pm$0.34}&\textbf{95.00$\pm$0.39}&\textbf{74.19$\pm$0.86}&\textbf{68.00$\pm$2.08}&\textbf{93.31$\pm$1.15}&\textbf{76.54$\pm$2.06}\\
\midrule
\multirow{14}{*}{60\%}&DCP-CV&TPAMI'22&85.85$\pm$0.91&65.00$\pm$3.38&32.98$\pm$1.05&91.72$\pm$0.29&60.01$\pm$0.83&56.67$\pm$1.09&29.19$\pm$2.71&46.54$\pm$2.63\\
&DCP-CG&TPAMI'22&82.95$\pm$0.46&66.19$\pm$3.96&32.40$\pm$0.91&91.16$\pm$0.36&61.73$\pm$0.79&52.69$\pm$0.55&31.91$\pm$2.29&51.03$\pm$1.84\\
&QMF&ICML'23&82.75$\pm$1.89&72.86$\pm$6.41&36.94$\pm$0.35&85.89$\pm$0.51&54.97$\pm$1.25&33.24$\pm$2.54&59.28$\pm$1.83&57.79$\pm$4.55\\
&PDF&ICML'24&83.17$\pm$1.01&77.14$\pm$7.08&36.27$\pm$0.52&87.58$\pm$0.65&54.87$\pm$1.88&33.79$\pm$2.70&60.28$\pm$2.20&57.06$\pm$5.08\\
&DUA-Nets$\bullet$&AAAI'21&74.35$\pm$1.79&63.10$\pm$5.95&19.17$\pm$0.43&76.24$\pm$0.56&39.03$\pm$1.34&30.98$\pm$2.20&50.72$\pm$2.40&33.38$\pm$2.79\\
&ETMC$\bullet$&TPAMI'22&80.15$\pm$1.95&72.14$\pm$3.54&34.45$\pm$0.46&83.04$\pm$0.69&56.58$\pm$1.41&33.21$\pm$2.35&65.75$\pm$1.52&58.60$\pm$3.01\\
&UIMC$\bullet$&CVPR'23&79.85$\pm$1.67&72.38$\pm$5.24&33.85$\pm$0.46&82.42$\pm$0.76&58.23$\pm$0.78&36.57$\pm$2.06&67.72$\pm$1.56&61.40$\pm$4.19\\
&ECML$\bullet$&AAAI'24&78.57$\pm$1.15&67.86$\pm$7.77&33.46$\pm$0.38&81.10$\pm$0.60&59.58$\pm$0.90&46.33$\pm$2.62&70.00$\pm$2.61&56.62$\pm$2.35\\
&TMNR$\bullet$&IJCAI'24&82.10$\pm$0.87&\underline{80.48$\pm$6.37}&19.15$\pm$0.32&83.77$\pm$0.70&51.25$\pm$1.43&35.90$\pm$2.35&51.22$\pm$2.55&65.22$\pm$4.01\\
&CCML$\bullet$&MM'24&82.02$\pm$1.77&74.52$\pm$7.23&30.07$\pm$0.33&81.87$\pm$0.62&54.23$\pm$1.17&31.02$\pm$1.75&64.41$\pm$2.49&61.54$\pm$4.80\\
&TUNED$\bullet$&AAAI'25&82.30$\pm$1.33&70.00$\pm$4.15&11.93$\pm$0.32&80.23$\pm$0.82&55.27$\pm$1.53&37.12$\pm$1.65&67.69$\pm$1.96&62.43$\pm$3.62\\
&SAEML$\bullet$&MM'25&84.60$\pm$1.16&80.00$\pm$5.02&34.00$\pm$0.63&83.10$\pm$0.67&63.89$\pm$0.73&56.05$\pm$1.77&72.00$\pm$1.83&63.24$\pm$3.45\\
&FUML$\bullet$&ICML'25&\underline{89.12$\pm$1.82}&78.33$\pm$7.93&\underline{37.16$\pm$0.41}&\underline{92.72$\pm$0.57}&\underline{69.30$\pm$0.94}&\underline{61.00$\pm$3.03}&\underline{88.09$\pm$1.38}&\underline{65.81$\pm$3.95}\\
&R-FUML$\bullet$&Ours&\textbf{92.32$\pm$1.32}&\textbf{80.95$\pm$7.30}&\textbf{37.70$\pm$0.49}&\textbf{92.89$\pm$0.58}&\textbf{71.40$\pm$0.71}&\textbf{64.33$\pm$2.02}&\textbf{89.84$\pm$1.51}&\textbf{66.32$\pm$4.01}\\
\bottomrule
\end{tabular}
}
\label{tab: VC results}
\end{center}
\vspace{-0.3cm}
\end{table*} 

\section{Experiments}
\subsection{Experimental Setup}
\subsubsection{Datasets}
For a thorough verification of the performance of our R-FUML, experiments are carried out on eight public datasets:
\textbf{Handwritten} (HW)\footnote{https://archive.ics.uci.edu/ml/datasets/Multiple+Features}, \textbf{MSRC-V1} (MSRC)~\cite{winn2005locus}, \textbf{NUS-WIDE-OBJ} (NUSOBJ)\footnote{https://lms.comp.nus.edu.sg/wp-content/uploads/2019/}, \textbf{Fashion-MV} (Fashion)~\cite{wang2023metaviewer}, \textbf{Scene15} (Scene)\footnote{https://doi.org/10.6084/m9.figshare.7007177.v1}, \textbf{LandUse}~\cite{yang2010bag}, \textbf{Leaves100} (Leaves)\footnote{https://archive.ics.uci.edu/dataset/241/one+hundred+plant+species+leaves+\\data+set}, and \textbf{PIE}\footnote{http://www.cs.cmu.edu/afs/cs/project/PIE/MultiPie/MultiPie/Home.html}. The training and the test sets are split in a ratio of 8:2. To create datasets with VC, we conduct two transformations: (1) In training and test sets, we alter the information in $V-2$ random views for different rates of multi-view instances, making these views' labels inconsistent with the true label, where $V$ refers to the number of views. (2) We add Gaussian noise with a standard deviation of 0.5 and a mean of 0.0 on a random number of modalities in $10\%$ of the test instances.

\subsubsection{Baselines}
For a comprehensive comparison, we adopted the following baselines: (1) Untrusted baselines, i.e., methods that do not provide decision uncertainty, include: \textbf{DCP}(CV\&CG)~\cite{lin2022dual}, \textbf{QMF}~\cite{zhang2023provable}, and \textbf{PDF}~\cite{cao2024predictive}. (2) The trusted baselines include: \textbf{DUA-Nets}~\cite{geng2021uncertainty}, \textbf{ETMC}~\cite{han2022trusted}, \textbf{UIMC}~\cite{xie2023exploring}, \textbf{ECML}~\cite{xu2024reliable}, \textbf{TMNR}~\cite{ijcai2024p582}, \textbf{CCML}~\cite{liu2024dynamic}, \textbf{TUNED}~\cite{huang2025trusted}, and \textbf{SAEML}~\cite{xu2025beyond}. Among them, \textbf{QMF}~\cite{zhang2023provable} and \textbf{PDF}~\cite{cao2024predictive} are designed for low-quality multi-view fusion. \textbf{TMNR}~\cite{ijcai2024p582} is designed for multi-view learning with label noise. For a fair comparison, we replace the backbone networks of \textbf{QMF}~\cite{zhang2023provable} and \textbf{PDF}~\cite{cao2024predictive} with fully connected layers while preserving their core models and loss functions. For the other baselines, we follow the settings in their source code.

\begin{figure*}[htbp]
    \centering
    
    \begin{subfigure}[b]{0.22\textwidth}
        \centering
        \includegraphics[width=\textwidth]{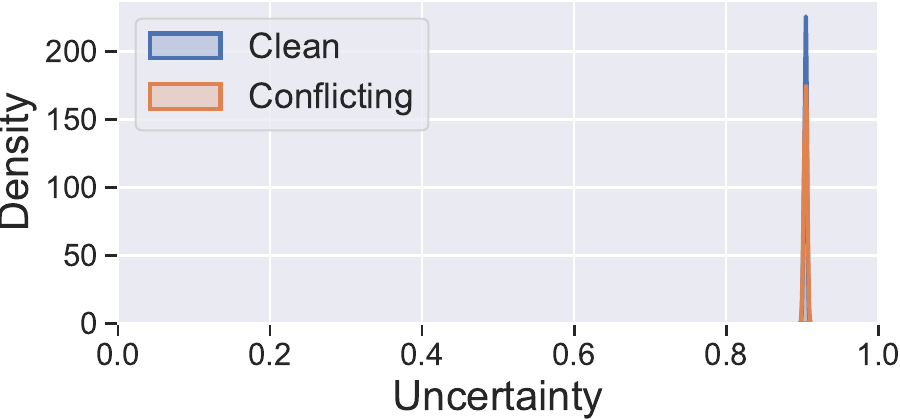}
        \caption{Initial distribution}
    \end{subfigure}
    \hfill
    \begin{subfigure}[b]{0.22\textwidth}
        \centering
        \includegraphics[width=\textwidth]{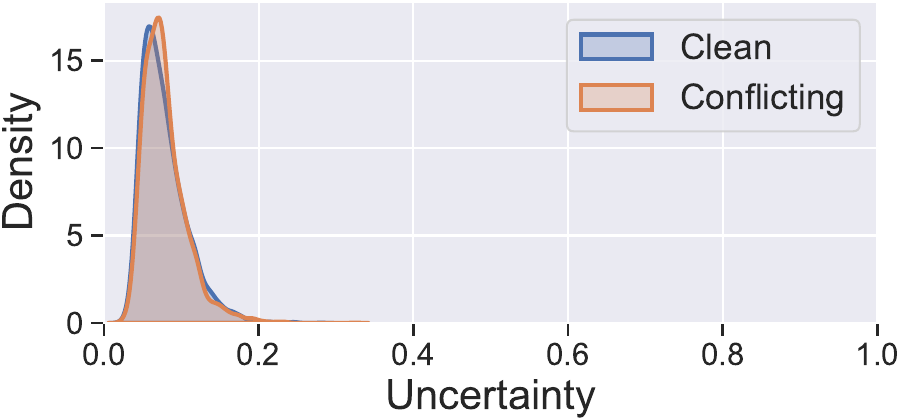}
        \caption{Epoch 10}
    \end{subfigure}
    \hfill
    \begin{subfigure}[b]{0.22\textwidth}
        \centering
        \includegraphics[width=\textwidth]{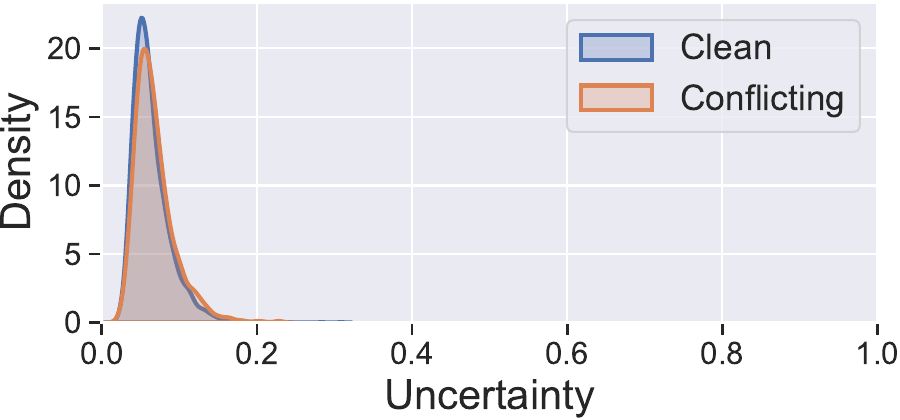}
        \caption{Epoch 50}
    \end{subfigure}
    \hfill
    \begin{subfigure}[b]{0.22\textwidth}
        \centering
        \includegraphics[width=\textwidth]{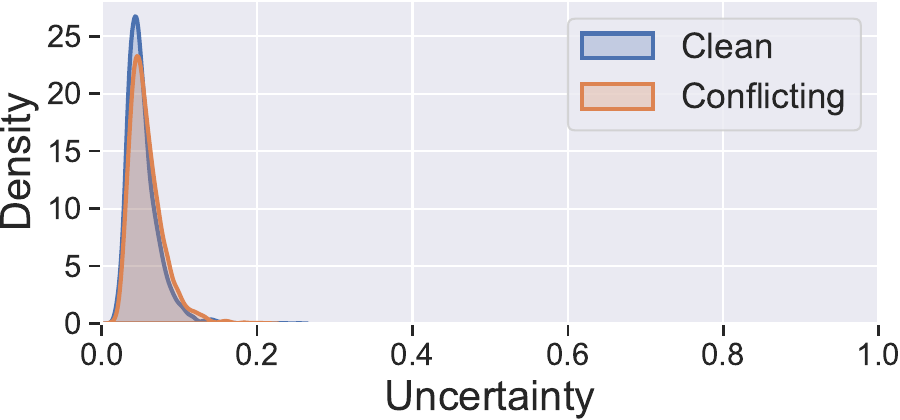}
        \caption{Epoch 500}
    \end{subfigure}
    \vspace{-0.2cm}
    \caption{Uncertainty density of clean and conflicting multi-view instances during training: SAEML's results on the Fashion dataset.}
    \label{fig:densityMap SAEML}
    \vspace{-0.2cm}
\end{figure*}

\begin{figure*}[htbp]
    \centering
    
    \begin{subfigure}[b]{0.22\textwidth}
        \centering
        \includegraphics[width=\textwidth]{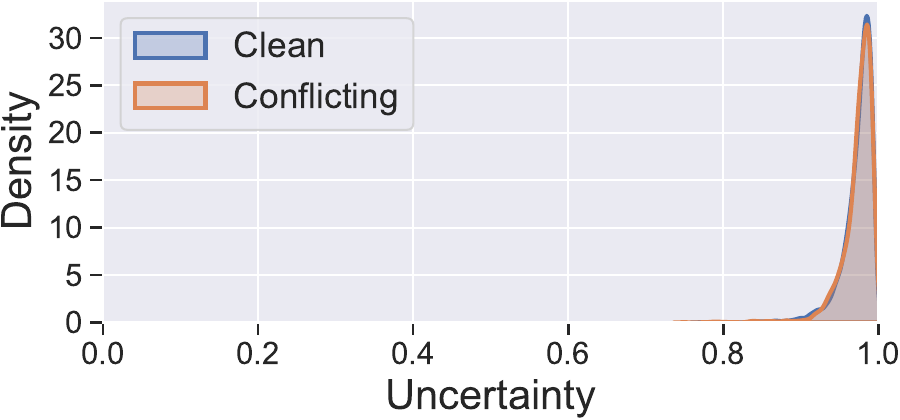}
        \caption{Initial distribution}
    \end{subfigure}
    \hfill
    \begin{subfigure}[b]{0.22\textwidth}
        \centering
        \includegraphics[width=\textwidth]{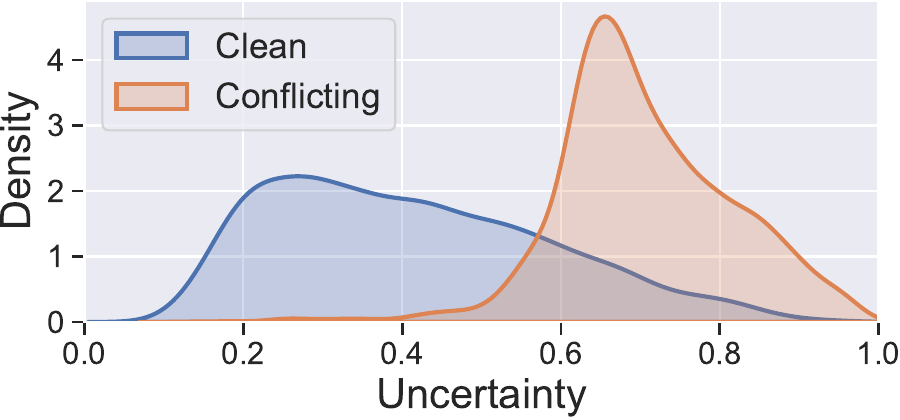}
        \caption{Epoch 10}
    \end{subfigure}
    \hfill
    \begin{subfigure}[b]{0.22\textwidth}
        \centering
        \includegraphics[width=\textwidth]{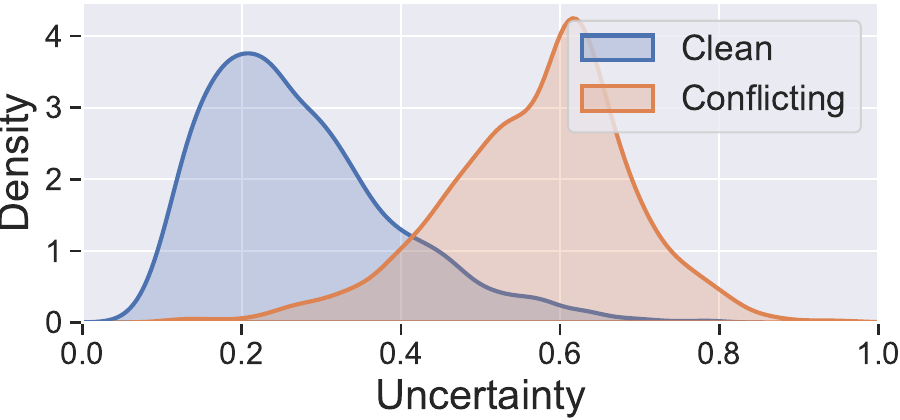}
        \caption{Epoch 50}
    \end{subfigure}
    \hfill
    \begin{subfigure}[b]{0.22\textwidth}
        \centering
        \includegraphics[width=\textwidth]{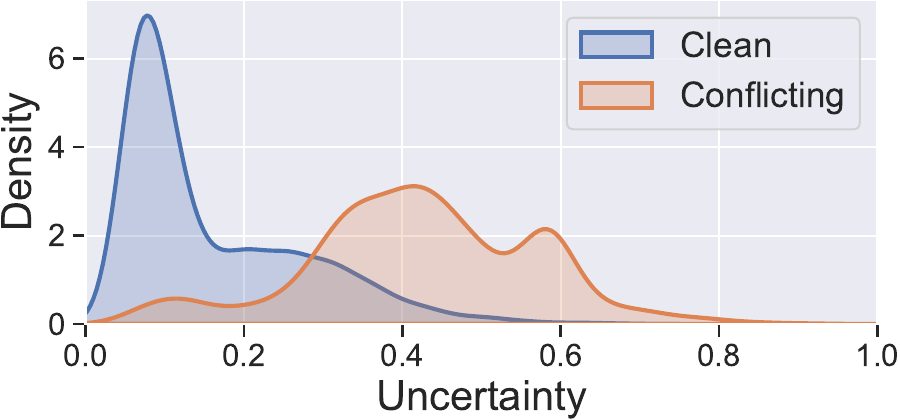}
        \caption{Epoch 500}
    \end{subfigure}
    \vspace{-0.2cm}
    \caption{Uncertainty density of clean and conflicting multi-view instances during training: FUML's results on the Fashion dataset.}
    \label{fig:densityMap FUML}
    \vspace{-0.2cm}

\end{figure*}

\begin{figure*}[htbp]
    \centering
    
    \begin{subfigure}[b]{0.22\textwidth}
        \centering
        \includegraphics[width=\textwidth]{imgs/Fashion_train_uncer_0_s.pdf}
        \caption{Initial distribution}
    \end{subfigure}
    \hfill
    \begin{subfigure}[b]{0.22\textwidth}
        \centering
        \includegraphics[width=\textwidth]{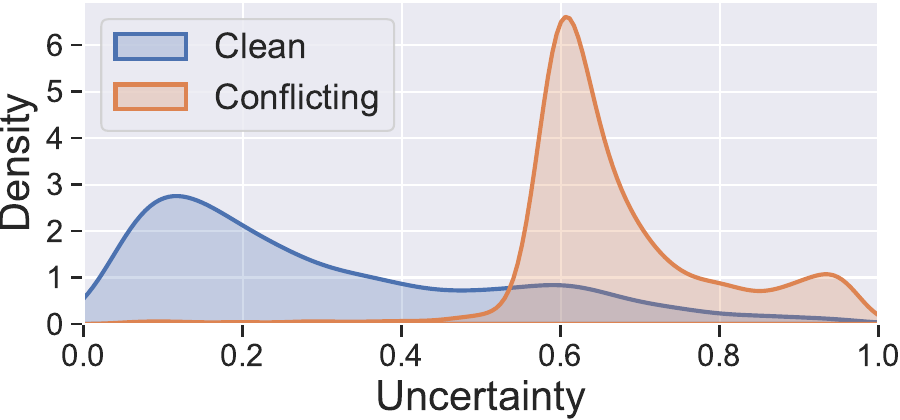}
        \caption{Epoch 10}
    \end{subfigure}
    \hfill
    \begin{subfigure}[b]{0.22\textwidth}
        \centering
        \includegraphics[width=\textwidth]{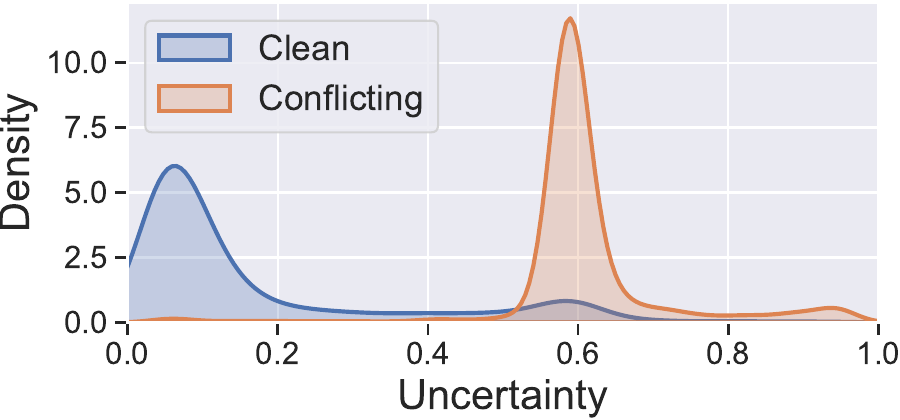}
        \caption{Epoch 50}
    \end{subfigure}
    \hfill
    \begin{subfigure}[b]{0.22\textwidth}
        \centering
        \includegraphics[width=\textwidth]{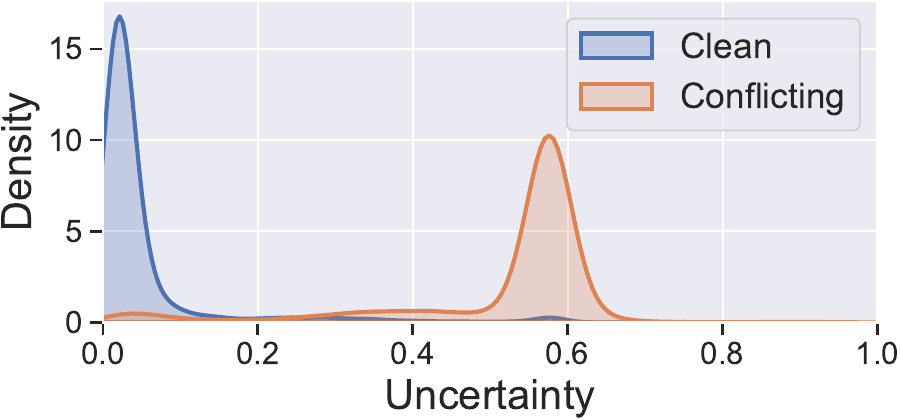}
        \caption{Epoch 500}
    \end{subfigure}
    \vspace{-0.2cm}
    \caption{Uncertainty density of clean and conflicting multi-view instances during training: R-FUML's results on the Fashion dataset.}
    \label{fig:densityMap R-FUML}
    \vspace{-0.2cm}

\end{figure*}

\begin{figure*}[htbp]
    \centering
    
    \begin{subfigure}[b]{0.22\textwidth}
        \centering
        \includegraphics[width=\textwidth]{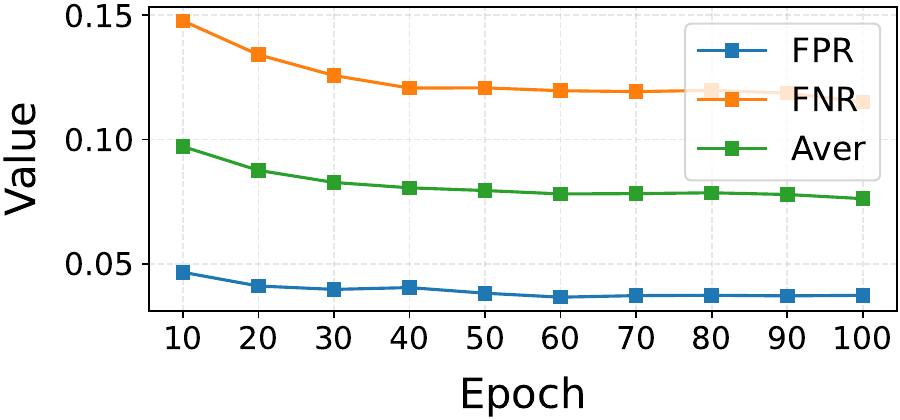}
        \caption{HW}
    \end{subfigure}
    \hfill
    \begin{subfigure}[b]{0.22\textwidth}
        \centering
        \includegraphics[width=\textwidth]{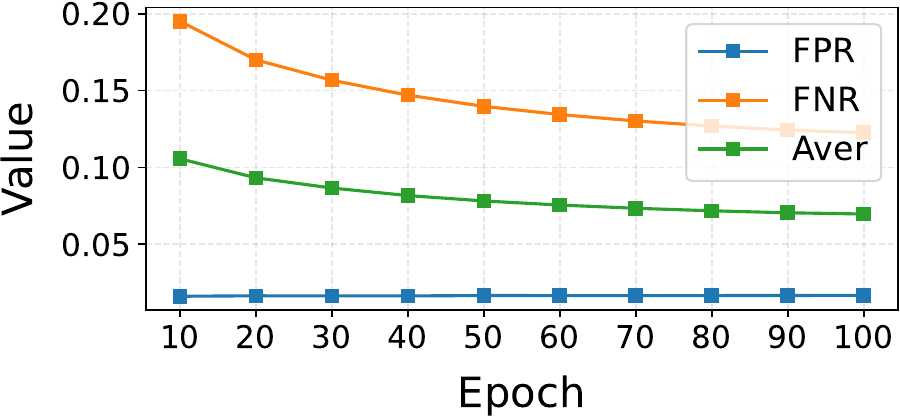}
        \caption{Fashion}
    \end{subfigure}
    \hfill
    \begin{subfigure}[b]{0.22\textwidth}
        \centering
        \includegraphics[width=\textwidth]{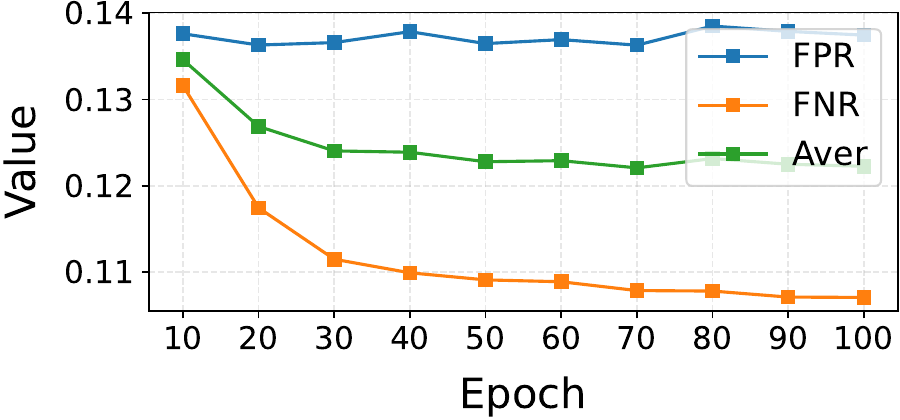}
        \caption{Scene}
    \end{subfigure}
    \hfill
    \begin{subfigure}[b]{0.22\textwidth}
        \centering
        \includegraphics[width=\textwidth]{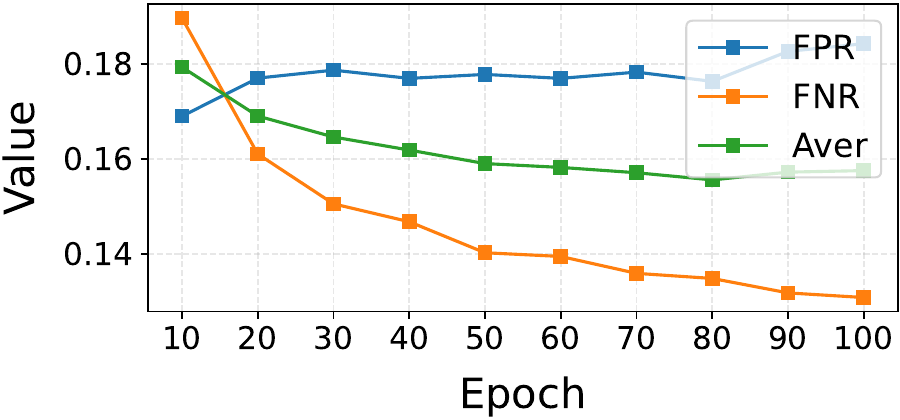}
        \caption{LandUse}
    \end{subfigure}

    \caption{R-FUML's ability to detect conflicting samples with different numbers of iterations on the HW, Fashion, Scene, and LandUse datasets. `Aver' denotes the average of FPR and FNR. 
    }
    \vspace{-0.2cm}
    \label{fig: pa cyc}
    \vspace{-0.2cm}
\end{figure*}

\subsubsection{Evaluation Metrics}

Given the randomness in the experimental setup, we report the mean accuracy and corresponding standard deviation (mean $\pm$ std) averaged over 10 independent runs with distinct random seeds, ensuring statistical reliability. We use the false positive rate at a true positive rate of $95\%$, i.e., FPR95, to quantitatively evaluate the proposed uncertainty estimation mechanism, where lower values indicate better performance.







\vspace{-0.2cm}
\subsection{Comparison with State-of-The-Arts}
To evaluate R-FUML, we perform multi-view classification on eight public datasets across 10 random seeds. The experimental results under different VC rates are shown in~\cref{tab: VC results}. The following observations can be made from these results: (1) As the VC rate increases from $0\%$ to $60\%$, the classification accuracy of all methods shows a gradual and remarkable decline due to misleading conflict information. For instance, on the HW dataset, DCP-CV, PDF, TUNED, SAEML, and FUML decrease by $12.90\%$, $15.23\%$, $15.98\%$, $14.27\%$, and $10.08\%$, respectively, whereas our R-FUML decreases by only $6.80\%$. This indicates the slower degradation and verifies its robustness, which is attributed to RMF and RLVC. (2) In general, as the VC rate increases, the advantage of our R-FUML method becomes more and more significant. When the VC rate reaches $60\%$, R-FUML outperforms the second-best methods (FUML and TMNR) on all eight datasets, with improvements such as $3.20\%$ on HW, $2.10\%$ on Scene, and $3.33\%$ on LandUse. This demonstrates the robustness of R-FUML in handling VC in both training and test sets. (3) R-FUML performs slightly worse than FUML on some datasets (such as NUSOBJ and PIE) under low or no VC, e.g., at VC rates of $0\%$ and $20\%$. This may be because the conflictive view division stage does not always accurately separate clean samples from conflicting samples. Although R-FUML slightly sacrifices minor accuracy at low VC rates, it provides more significant performance gains at medium and high VC rates, which is a reasonable trade-off for practical scenarios.

\begin{table}[tbp]
\caption{Quantitative uncertainty estimation results on in-distribution (clean) vs. out-of-distribution (conflicting) samples, measured by FPR95.}
\vspace{-0.2cm}
\scriptsize
\renewcommand\arraystretch{0.72} 
\tabcolsep=0.33cm
\begin{center}
\begin{tabular}{c| c | c |c |c }
\toprule[0.3mm]
\diagbox{Dataset}{Method} & TUNED&SAEML&FUML&R-FUML\\
\toprule
HW& 0.852 & 0.874 &0.518 &\textbf{0.218}\\
Fashion& 0.880 & 0.923 & 0.410 & \textbf{0.086}\\
Scene& 0.940 & 0.934&0.604 &\textbf{0.490}\\
LandUse&0.937 &0.937&0.703&\textbf{0.573}\\
\bottomrule
\end{tabular}
\label{Tab: Quantitative results of uncertainty}
\vspace{-0.3cm}
\end{center}
\end{table} 

\begin{table}[tbp]
\caption{Ablation study of model components with a $40\%$ VC in the training and test sets. `DRF' is from~\cite{duandeep}. The evaluation metric is accuracy(\%), reported as mean$\pm$std. }
\vspace{-0.2cm}
\scriptsize
\renewcommand\arraystretch{0.72} 

\tabcolsep=0.16cm
\begin{center}
\begin{tabular}{c c c | c c c c }
\toprule
\multicolumn{3}{c|}{Method variants}&\multicolumn{4}{c}{Dataset}\\
\midrule
$\mathcal{L}^a_{ccl}$&$\mathcal{L}^v_{ccl}$& Rule & HW & Fashion &Scene & LandUse\\
\midrule
$\surd$&$\times$& Concat &86.22$\pm$5.22&90.06$\pm$0.48&67.09$\pm$0.90&57.67$\pm$1.83\\
$\times$&$\surd$& Avg &95.30$\pm$1.26&94.14$\pm$0.45&73.95$\pm$1.15&67.19$\pm$1.67\\
$\surd$&$\times$& Avg &94.05$\pm$1.42&94.69$\pm$0.38&73.87$\pm$0.58&66.48$\pm$1.80\\
$\surd$&$\surd$& Avg &94.62$\pm$1.06&94.40$\pm$0.45&74.02$\pm$0.82&67.83$\pm$2.26\\
$\times$&$\surd$& DRF &94.58$\pm$1.43&94.89$\pm$0.56&74.11$\pm$0.69&67.36$\pm$2.06\\
$\surd$&$\times$& DRF &\textbf{95.35$\pm$1.06}&94.26$\pm$0.60&73.96$\pm$0.98&66.74$\pm$2.10\\
$\surd$&$\surd$& DRF&\textbf{95.35}$\pm$1.26&94.78$\pm$0.49&73.98$\pm$0.93&67.71$\pm$2.45\\
$\times$&$\surd$& RMF &\textbf{95.35$\pm$1.17}&94.24$\pm$0.41&73.96$\pm$0.98&67.36$\pm$2.06\\
$\surd$&$\times$& RMF &93.43$\pm$1.23&94.77$\pm$0.43&73.86$\pm$0.76&66.71$\pm$2.19\\
\midrule
$\surd$&$\surd$& RMF &\textbf{95.35$\pm$1.22}&\textbf{95.00$\pm$0.39}&\textbf{74.19$\pm$0.86}&\textbf{68.00$\pm$2.08}\\
\bottomrule
\end{tabular}
\label{Tab:Ab}
\end{center}
\vspace{-0.5cm}
\end{table}

\subsection{Robustness Study}

In this section, we present visualization results of multi-view training to comprehensively evaluate the robustness and effectiveness of our R-FUML. For fair comparison, R-FUML is benchmarked against state-of-the-art baselines using `Stage 4: Robust Learning'. As shown in~\Cref{fig:robustness study}, the following observation can be made: (1) Baseline methods, including FUML, TMNR, and DCP-CG, exhibit a pronounced performance drop in the later Stages of training. This degradation indicates that these methods gradually overfit conflicting samples, undermining their generalization ability. In contrast, our R-FUML (red curve) maintains stable and high performance throughout the entire training process, with no observable drop-off. (2) R-FUML consistently outperforms all competing methods across almost all epochs on each dataset. These results demonstrate that our R-FUML not only achieves strong multi-view classification performance but also effectively alleviates overfitting to conflicting samples, thereby enhancing the reliability and generalization of multi-view learning.



 
\subsection{Uncertainty Effectiveness Analysis}

\subsubsection{Qualitative Results}
To qualitatively validate the effectiveness of our R-FUML in estimating uncertainty for multi-view instances with VC, i.e., conflicting multi-view instances, we compare it with two recent EDL-based TMVC methods, i.e., TUNED and SAEML, and a Fuzzy-based TMVC method (FUML), using uncertainty density maps. The results are illustrated in~\Cref{fig:densityMap SAEML},~\Cref{fig:densityMap FUML}, and~\Cref{fig:densityMap R-FUML}, respectively. These results show the following: 
(1) As the number of iterations increases, the uncertainty of SAEML's estimation of training instances gradually decreases. However, SAEML cannot distinguish clean instances from conflicting instances, i.e., instances with VC. This is partly because its uncertainty quantification mechanism neglects the inherent conflict between belief masses, and partly because these methods lack a view-conflict detection mechanism, tend to overfit training instances with VC, and ultimately yield inaccurate uncertainty quantification.
(2) For FUML, as training progresses, the estimated uncertainty of clean instances gradually decreases, and the model gradually becomes able to distinguish clean instances from instances with VC. However, even at the end of training (epoch=500), there is still a significant overlap between the clean sample and the conflicting sample regions. This is because FUML lacks a VC detection mechanism, leading to overfitting on conflicting instances. 
(3) For our R-FUML, as the number of epochs increases, the uncertainty of clean instances gradually decreases, and the overlap between clean and conflicting instances is gradually reduced. At the end of training, clean and conflicting instances have almost no overlap, as shown in the right part of \Cref{method1}. These results demonstrate that R-FUML can accurately measure the uncertainty of conflicting multi-view instances, owing to the proposed fuzzy-based uncertainty estimation mechanism and RLVC framework.

\begin{table}[tbp]
\caption{Ablation study of training stages with $40\%$ VC in the training and test sets. 
Metrics are reported as mean$\pm$std of accuracy (\%). 
`Pre.' denotes Pre-training; `Cyc.' denotes Cyclical Training. 
`$\times$' for `Cyc.' indicates a fixed learning rate in `stage~2' (non-periodic).}
\vspace{-0.2cm}
\scriptsize
\renewcommand\arraystretch{0.72} 
\tabcolsep=0.23cm
\begin{center}
\begin{tabular}{c c | c c c c }
\toprule
\multicolumn{2}{c|}{Method variants}&\multicolumn{4}{c}{Dataset}\\
\midrule
Pre.& Cyc. & HW & Fashion &Scene & LandUse\\
\midrule
$\times$&$\times$&93.60$\pm$1.12&94.14$\pm$0.39&73.86$\pm$0.82&66.40$\pm$1.57\\
$\surd$&$\times$&94.68$\pm$1.28&94.22$\pm$0.36&73.69$\pm$0.66&67.05$\pm$1.93\\
$\times$&$\surd$&95.00$\pm$1.11&94.84$\pm$0.45&74.15$\pm$0.72&67.00$\pm$1.54\\
\midrule
$\surd$&$\surd$&\textbf{95.35$\pm$1.22}&\textbf{95.00$\pm$0.39}&\textbf{74.19$\pm$0.86}&\textbf{68.00$\pm$2.08}\\
\bottomrule
\end{tabular}
\label{Tab:Ab2}
\end{center}
\vspace{-0.4cm}
\end{table}

\subsubsection{Quantitative Results}
We further conduct a quantitative evaluation using FPR95 to validate the effectiveness of the uncertainty estimation mechanism of our R-FUML on the HW, Fashion, Scene, and LandUse datasets. 
As shown in Table~\ref{Tab: Quantitative results of uncertainty}, TUNED and SAEML yield high FPR95 values above 0.85 on most datasets, indicating limited ability to distinguish clean from conflicting multi-view instances. FUML achieves moderate improvement but still performs unsatisfactorily on complex datasets.
In contrast, our R-FUML consistently achieves the lowest FPR95 across all datasets and remarkably outperforms all baselines. These quantitative results confirm that R-FUML enables more accurate uncertainty estimation for multi-view data with VC.

\subsection{Ablation Study}

To validate the effectiveness of each component in our proposed R-FUML, we conduct ablation studies on the HW, Fashion, Scene, and LandUse datasets, with the VC rate fixed at $40\%$ in
both the training and test sets. We conduct experiments over 10 random seeds and report the mean values.

\textbf{First}, we ablate each component of the loss function and the multi-view fusion strategy.
For simplicity, in this section, we denotes $\mathcal{L}_{ccl}(\mathbf{r}^{a},\mathbf{y})$ as $\mathcal{L}_{ccl}^a$ and represent $\sum_{v=1}^V \mathcal{L}_{ccl}(\mathbf{r}^{v}, \mathbf{y})$ as $\mathcal{L}_{ccl}^v$. The results, presented in \Cref{Tab:Ab} show the following: (1) Removing either $\mathcal{L}_{ccl}^a$ or $\mathcal{L}_{ccl}^v$ degrades performance to varying degrees, indicating that both components in the total loss function are necessary. (2) Feature-level fusion, i.e., concatenating (Concat) all features and only using a single DNN for prediction, causes a substantial performance decline. (3) Compared with the arithmetic mean (Avg) and double reliable multi-view fusion (DRF)~\cite{duandeep}, our RMF achieves superior performance. This is because RMF effectively mitigates the impact of VC during both training and testing.

\textbf{Second}, to evaluate the effectiveness of the pre-training and cyclical training stages in the proposed robust learning pipeline, we conduct additional ablation studies. As shown in \Cref{Tab:Ab2}, both stages are critical for learning under VC, and removing either stage results in clear performance degradation. On one hand, removing `Stage 1: Pre-training' means that the loss is computed from the initial training iteration, making it difficult to effectively distinguish between clean samples and conflicting samples in early training. On the other hand, removing `Stage 2: Cyclical Training' could cause the model to overfit and fail to fully exploit the memory effect of deep neural networks.







\subsection{Parameter Analysis}



In this section, we explore the impact of the cycle count in our R-FUML. This experiment is conducted with a $40\%$ VC rate on the HW, Fashion, Scene, and LandUse datasets, and the cycle length is fixed at 10 epochs. \Cref{fig: pa cyc} shows the average FNR (false negative rate) and FPR (false positive rate) across all views using different cycle counts. We can observe the following. (1) As the cycle count increases, i.e., as the total number of epochs increases, the FNR decreases and then stabilizes across all datasets. This indicates that increasing the cycle count effectively prevents the model from classifying conflicting samples as clean samples. (2) As the cycle count increases, on the HW and Fashion datasets initially decreases and then stabilizes. This indicates that increasing the cycle count effectively prevents the model from classifying clean views as conflicting views. In contrast, on the Scene dataset, the FPR does not decrease; even on the LandUse dataset, the FPR slightly increased. This suggests that a larger cycle count is not always perferable. (3) Across all datasets, as the cycle count increases, the average of FNR and FNR initially decreases and then tends to stabilize. This indicates that the overall capability of VC detection improves with increasing cycle count. Considering all datasets, we recommend setting the cycle count to 10, corresponding to 100 total epochs in `Stage 2: Cyclical Training'.




\section{Conclusion}

In this paper, we reveal and investigate a challenging problem for trusted multi-view classification: how to mitigate the impact of view conflict (VC) during both training and testing.
To address this issue, we propose a \underline{R}obust \underline{F}uzzy \underline{M}ulti-view \underline{L}earning (R-FUML) framework. Grounded in Fuzzy Set Theory, R-FUML models category credibility and quantifies uncertainty via entropy, incorporates Robust Multi-view Fusion (RMF) to mitigate the impact of conflicting views, and leverages Robust Learning against VC (RLVC) for robust training. Extensive experiments on eight public datasets demonstrate that R-FUML outperforms 15 state-of-the-art baselines, especially under high VC rates, with superior robustness and accurate uncertainty estimation.



\section*{Acknowledgment}
This work was supported in part by the National Natural Science Foundation of China under Grant U25A20534, 62472295, 62372315, and U25B6003; in part by the Fundamental Research Funds for the Central Universities under Grant CJ202303 and CJ202403; in part by Sichuan Science and Technology Planning Project under Grant 2024NSFTD0049, and 24NSFTD0130; in part by the Central Government's Guide to Local Science and Technology Development Fund under Grant 2025ZYDF101; in part by the Chengdu Science and Technology Project Grants No. 2025-YF08-00104-GX and 2025-YF05-00169-SN; and in part by the Luzhou City School-Local-Enterprise-Academy Science and Technology Cooperation Project under Grant 2024XDY200.



%





\ifCLASSOPTIONcaptionsoff
  \newpage
\fi





\bibliographystyle{IEEEtran}
\bibliography{Bibliography}

\begin{thebibliography}{10}
\providecommand{\url}[1]{#1}
\csname url@rmstyle\endcsname
\providecommand{\newblock}{\relax}
\providecommand{\bibinfo}[2]{#2}
\providecommand\BIBentrySTDinterwordspacing{\spaceskip=0pt\relax}
\providecommand\BIBentryALTinterwordstretchfactor{4}
\providecommand\BIBentryALTinterwordspacing{\spaceskip=\fontdimen2\font plus
\BIBentryALTinterwordstretchfactor\fontdimen3\font minus
  \fontdimen4\font\relax}
\providecommand\BIBforeignlanguage[2]{{%
\expandafter\ifx\csname l@#1\endcsname\relax
\typeout{** WARNING: IEEEtran.bst: No hyphenation pattern has been}%
\typeout{** loaded for the language `#1'. Using the pattern for}%
\typeout{** the default language instead.}%
\else
\language=\csname l@#1\endcsname
\fi
#2}}

\bibitem{guo2020deep}
Y.~Guo, H.~Wang, Q.~Hu, H.~Liu, L.~Liu, and M.~Bennamoun, ``Deep learning for
  3d point clouds: A survey,'' \emph{IEEE transactions on pattern analysis and
  machine intelligence}, vol.~43, no.~12, pp. 4338--4364, 2020.

\bibitem{feng2026robust}
Y.~Feng, H.~Zhu, D.~Peng, X.~Peng, X.~Song, and P.~Hu, ``Robust cross-modal
  alignment learning for cross-scene spatial reasoning and grounding,''
  \emph{Advances in Neural Information Processing Systems}, vol.~38, pp.
  49\,055--49\,075, 2026.

\bibitem{zhang2024multimodal}
Q.~Zhang, Y.~Wei, Z.~Han, H.~Fu, X.~Peng, Q.~Hu, C.~Deng, C.~Xu, J.~Wen, D.~Hu,
  \emph{et~al.}, ``Multimodal fusion on low-quality data: A comprehensive
  survey,'' \emph{Information Fusion}, p. 104437, 2026.

\bibitem{wang2024real}
C.~Wang, W.~Zhu, B.-B. Gao, Z.~Gan, J.~Zhang, Z.~Gu, S.~Qian, M.~Chen, and
  L.~Ma, ``Real-iad: A real-world multi-view dataset for benchmarking versatile
  industrial anomaly detection,'' in \emph{Proceedings of the IEEE/CVF
  Conference on Computer Vision and Pattern Recognition}, 2024, pp.
  22\,883--22\,892.

\bibitem{gong2025carp}
Z.~Gong, P.~Ding, S.~Lyu, S.~Huang, M.~Sun, W.~Zhao, Z.~Fan, and D.~Wang,
  ``Carp: Visuomotor policy learning via coarse-to-fine autoregressive
  prediction,'' in \emph{Proceedings of the IEEE/CVF International Conference
  on Computer Vision}, 2025, pp. 13\,460--13\,470.

\bibitem{yadav2023deep}
A.~Yadav and D.~K. Vishwakarma, ``A deep multi-level attentive network for
  multimodal sentiment analysis,'' \emph{ACM Transactions on Multimedia
  Computing, Communications and Applications}, vol.~19, no.~1, pp. 1--19, 2023.

\bibitem{xu2025distilled}
Y.~Xu, F.~Zhou, C.~Zhao, Y.~Wang, C.~Yang, and H.~Chen, ``Distilled prompt
  learning for incomplete multimodal survival prediction,'' in
  \emph{Proceedings of the Computer Vision and Pattern Recognition Conference},
  2025, pp. 5102--5111.

\bibitem{han2022trusted}
Z.~Han, C.~Zhang, H.~Fu, and J.~T. Zhou, ``Trusted multi-view classification
  with dynamic evidential fusion,'' \emph{IEEE Transactions on Pattern Analysis
  and Machine Intelligence}, vol.~45, no.~2, pp. 2551--2566, 2022.

\bibitem{sensoy2018evidential}
M.~Sensoy, L.~Kaplan, and M.~Kandemir, ``Evidential deep learning to quantify
  classification uncertainty,'' \emph{Advances in Neural Information Processing
  Systems}, vol.~31, 2018.

\bibitem{shafer1992dempster}
G.~Shafer, ``Dempster-shafer theory,'' \emph{Encyclopedia of artificial
  intelligence}, vol.~1, pp. 330--331, 1992.

\bibitem{liu2022trusted}
W.~Liu, X.~Yue, Y.~Chen, and T.~Denoeux, ``Trusted multi-view deep learning
  with opinion aggregation,'' in \emph{Proceedings of the AAAI Conference on
  Artificial Intelligence}, vol.~36, no.~7, 2022, pp. 7585--7593.

\bibitem{liu2024dynamic}
Y.~Liu, L.~Liu, C.~Xu, X.~Song, Z.~Guan, and W.~Zhao, ``Dynamic evidence
  decoupling for trusted multi-view learning,'' in \emph{Proceedings of the
  32nd ACM International Conference on Multimedia}, 2024, pp. 7269--7277.

\bibitem{liu2025enhancing}
W.~Liu, Y.~Chen, and X.~Yue, ``Enhancing multi-view classification reliability
  with adaptive rejection,'' in \emph{Proceedings of the AAAI Conference on
  Artificial Intelligence}, vol.~39, no.~18, 2025, pp. 18\,969--18\,977.

\bibitem{xie2023exploring}
M.~Xie, Z.~Han, C.~Zhang, Y.~Bai, and Q.~Hu, ``Exploring and exploiting
  uncertainty for incomplete multi-view classification,'' in \emph{Proceedings
  of the IEEE/CVF Conference on Computer Vision and Pattern Recognition}, 2023,
  pp. 19\,873--19\,882.

\bibitem{ijcai2024p582}
C.~Xu, Y.~Zhang, Z.~Guan, and W.~Zhao, ``Trusted multi-view learning with label
  noise,'' in \emph{Proceedings of the Thirty-Third International Joint
  Conference on Artificial Intelligence, {IJCAI-24}}, K.~Larson, Ed.\hskip 1em
  plus 0.5em minus 0.4em\relax International Joint Conferences on Artificial
  Intelligence Organization, 8 2024, pp. 5263--5271, main Track.

\bibitem{wang2025reliable}
X.~Wang, S.~Duan, Q.~Li, G.~Duan, Y.~Sun, and D.~Peng, ``Reliable
  disentanglement multi-view learning against view adversarial attacks,''
  \emph{arXiv preprint arXiv:2505.04046}, 2025.

\bibitem{huang2024noise}
Z.~Huang, M.~Yang, X.~Xiao, P.~Hu, and X.~Peng, ``Noise-robust vision-language
  pre-training with positive-negative learning,'' \emph{IEEE Transactions on
  Pattern Analysis and Machine Intelligence}, 2024.

\bibitem{lai2025rethinking}
H.~Lai, G.~Xiong, H.~Mai, X.~Liu, and T.~Zhang, ``Rethinking noisy video-text
  retrieval via relation-aware alignment,'' in \emph{Proceedings of the
  Computer Vision and Pattern Recognition Conference}, 2025, pp. 9231--9241.

\bibitem{xiao2019multi}
F.~Xiao, ``Multi-sensor data fusion based on the belief divergence measure of
  evidences and the belief entropy,'' \emph{Information Fusion}, vol.~46, pp.
  23--32, 2019.

\bibitem{shang2021compound}
Q.~Shang, H.~Li, Y.~Deng, and K.~H. Cheong, ``Compound credibility for
  conflicting evidence combination: An autoencoder-k-means approach,''
  \emph{IEEE Transactions on Systems, Man, and Cybernetics: Systems}, vol.~52,
  no.~9, pp. 5602--5610, 2021.

\bibitem{zadeh1965fuzzy}
L.~A. Zadeh, ``Fuzzy sets,'' \emph{Information and Control}, vol.~8, no.~3, pp.
  338--353, 1965.

\bibitem{liu2025reliable}
C.~Liu, J.~Wen, Y.~Xu, B.~Zhang, L.~Nie, and M.~Zhang, ``Reliable
  representation learning for incomplete multi-view missing multi-label
  classification,'' \emph{IEEE Transactions on Pattern Analysis and Machine
  Intelligence}, 2025.

\bibitem{wen2026multi}
J.~Wen, J.~Zhou, X.~Lu, Y.~Tian, Z.~Zhang, L.~Shen, and Y.~Xu, ``Multi-domain
  feature integration based trusted partial multi-view incomplete multi-label
  learning,'' \emph{IEEE Transactions on Pattern Analysis and Machine
  Intelligence}, 2026.

\bibitem{chaudhuri2009multi}
K.~Chaudhuri, S.~M. Kakade, K.~Livescu, and K.~Sridharan, ``Multi-view
  clustering via canonical correlation analysis,'' in \emph{Proceedings of the
  26th Annual International Conference on Machine Learning}, 2009, pp.
  129--136.

\bibitem{rupnik2010multi}
J.~Rupnik and J.~Shawe-Taylor, ``Multi-view canonical correlation analysis,''
  in \emph{Conference on Data Mining and Data Warehouses (SiKDD 2010)}, vol.
  473, 2010, pp. 1--4.

\bibitem{yan2020higcin}
R.~Yan, L.~Xie, J.~Tang, X.~Shu, and Q.~Tian, ``Higcin: Hierarchical
  graph-based cross inference network for group activity recognition,''
  \emph{IEEE Transactions on Pattern Analysis and Machine Intelligence},
  vol.~45, no.~6, pp. 6955--6968, 2020.

\bibitem{cao2024predictive}
B.~Cao, Y.~Xia, Y.~Ding, C.~Zhang, and Q.~Hu, ``Predictive dynamic fusion,''
  \emph{International Conference on Machine Learning}, 2024.

\bibitem{bi2024sample}
J.~Bi and F.~Dornaika, ``Sample-weighted fused graph-based semi-supervised
  learning on multi-view data,'' \emph{Information Fusion}, vol. 104, p.
  102175, 2024.

\bibitem{zhan2025elip}
G.~Zhan, Y.~Liu, K.~Han, W.~Xie, and A.~Zisserman, ``Elip: Enhanced
  visual-language foundation models for image retrieval,'' in \emph{2025
  International Conference on Content-Based Multimedia Indexing (CBMI)}.\hskip
  1em plus 0.5em minus 0.4em\relax IEEE, 2025, pp. 1--8.

\bibitem{han2020trusted}
Z.~Han, C.~Zhang, H.~Fu, and J.~T. Zhou, ``Trusted multi-view classification,''
  in \emph{International Conference on Learning Representations}, 2020.

\bibitem{xu2024reliable}
C.~Xu, J.~Si, Z.~Guan, W.~Zhao, Y.~Wu, and X.~Gao, ``Reliable conflictive
  multi-view learning,'' in \emph{Proceedings of the AAAI Conference on
  Artificial Intelligence}, vol.~38, no.~14, 2024, pp. 16\,129--16\,137.

\bibitem{yue2025evidential}
X.~Yue, Z.~Dong, Y.~Chen, and S.~Xie, ``Evidential dissonance measure in robust
  multi-view classification to resist adversarial attack,'' \emph{Information
  Fusion}, vol. 113, p. 102605, 2025.

\bibitem{chen2023uncertainty}
M.~Chen, J.~Gao, and C.~Xu, ``Uncertainty-aware dual-evidential learning for
  weakly-supervised temporal action localization,'' \emph{IEEE Transactions on
  Pattern Analysis and Machine Intelligence}, 2023.

\bibitem{wang2020survey}
H.~Wang and D.-Y. Yeung, ``A survey on bayesian deep learning,'' \emph{ACM
  computing surveys (csur)}, vol.~53, no.~5, pp. 1--37, 2020.

\bibitem{gal2016dropout}
Y.~Gal and Z.~Ghahramani, ``Dropout as a bayesian approximation: Representing
  model uncertainty in deep learning,'' in \emph{international conference on
  machine learning}.\hskip 1em plus 0.5em minus 0.4em\relax PMLR, 2016, pp.
  1050--1059.

\bibitem{lakshminarayanan2017simple}
B.~Lakshminarayanan, A.~Pritzel, and C.~Blundell, ``Simple and scalable
  predictive uncertainty estimation using deep ensembles,'' \emph{Advances in
  Neural Information Processing Systems}, vol.~30, 2017.

\bibitem{lyzhov2020greedy}
A.~Lyzhov, Y.~Molchanova, A.~Ashukha, D.~Molchanov, and D.~Vetrov, ``Greedy
  policy search: A simple baseline for learnable test-time augmentation,'' in
  \emph{Conference on Uncertainty in Artificial Intelligence}.\hskip 1em plus
  0.5em minus 0.4em\relax PMLR, 2020, pp. 1308--1317.

\bibitem{gao2025comprehensive}
J.~Gao, M.~Chen, L.~Xiang, and C.~Xu, ``A comprehensive survey on evidential
  deep learning and its applications,'' \emph{IEEE Transactions on Pattern
  Analysis and Machine Intelligence}, 2025.

\bibitem{li2025deep}
Y.~Li, L.~Zhen, Y.~Sun, D.~Peng, X.~Peng, and P.~Hu, ``Deep evidential hashing
  for trustworthy cross-modal retrieval,'' in \emph{Proceedings of the AAAI
  Conference on Artificial Intelligence}, vol.~39, no.~17, 2025, pp.
  18\,566--18\,574.

\bibitem{huang2021learning}
Z.~Huang, G.~Niu, X.~Liu, W.~Ding, X.~Xiao, H.~Wu, and X.~Peng, ``Learning with
  noisy correspondence for cross-modal matching,'' \emph{Advances in Neural
  Information Processing Systems}, vol.~34, pp. 29\,406--29\,419, 2021.

\bibitem{hu2023cross}
P.~Hu, Z.~Huang, D.~Peng, X.~Wang, and X.~Peng, ``Cross-modal retrieval with
  partially mismatched pairs,'' \emph{IEEE Transactions on Pattern Analysis and
  Machine Intelligence}, vol.~45, no.~8, pp. 9595--9610, 2023.

\bibitem{wang2025noisy}
Y.~Wang, Y.~Wu, Z.~Dai, C.~Tian, J.~Long, and J.~Chen, ``Noisy correspondence
  rectification via asymmetric similarity learning,'' in \emph{Proceedings of
  the AAAI Conference on Artificial Intelligence}, vol.~39, no.~20, 2025, pp.
  21\,384--21\,392.

\bibitem{qin2024noisy}
Y.~Qin, Y.~Chen, D.~Peng, X.~Peng, J.~T. Zhou, and P.~Hu,
  ``Noisy-correspondence learning for text-to-image person re-identification,''
  in \emph{Proceedings of the IEEE/CVF Conference on Computer Vision and
  Pattern Recognition}, 2024, pp. 27\,197--27\,206.

\bibitem{sun2025roll}
Y.~Sun, Y.~Li, Z.~Ren, G.~Duan, D.~Peng, and P.~Hu, ``Roll: Robust noisy
  pseudo-label learning for multi-view clustering with noisy correspondence,''
  in \emph{Proceedings of the Computer Vision and Pattern Recognition
  Conference}, 2025, pp. 30\,732--30\,741.

\bibitem{han2024noise}
H.~Han, Q.~Zheng, M.~Luo, K.~Miao, F.~Tian, and Y.~Chen, ``Noise-tolerant
  learning for audio-visual action recognition,'' \emph{IEEE Transactions on
  Multimedia}, vol.~26, pp. 7761--7774, 2024.

\bibitem{lin2024multi}
Y.~Lin, J.~Zhang, Z.~Huang, J.~Liu, Z.~Wen, and X.~Peng, ``Multi-granularity
  correspondence learning from long-term noisy videos,'' \emph{arXiv preprint
  arXiv:2401.16702}, 2024.

\bibitem{lin2023graph}
Y.~Lin, M.~Yang, J.~Yu, P.~Hu, C.~Zhang, and X.~Peng, ``Graph matching with
  bi-level noisy correspondence,'' in \emph{Proceedings of the IEEE/CVF
  international conference on computer vision}, 2023, pp. 23\,362--23\,371.

\bibitem{sun2024robust}
Y.~Sun, Y.~Qin, Y.~Li, D.~Peng, X.~Peng, and P.~Hu, ``Robust multi-view
  clustering with noisy correspondence,'' \emph{IEEE Transactions on Knowledge
  and Data Engineering}, 2024.

\bibitem{liu2010uncertainty}
B.~Liu and B.~Liu, \emph{Uncertainty theory}.\hskip 1em plus 0.5em minus
  0.4em\relax Springer, 2010.

\bibitem{duan2025fuzzy}
S.~Duan, Y.~Sun, D.~Peng, Z.~Liu, X.~Song, and P.~Hu, ``Fuzzy multimodal
  learning for trusted cross-modal retrieval,'' in \emph{Proceedings of the
  Computer Vision and Pattern Recognition Conference}, 2025, pp.
  20\,747--20\,756.

\bibitem{li2020dividemix}
J.~Li, R.~Socher, and S.~C. Hoi, ``Dividemix: Learning with noisy labels as
  semi-supervised learning,'' \emph{arXiv preprint arXiv:2002.07394}, 2020.

\bibitem{yang2024robust}
M.~Yang, Z.~Huang, and X.~Peng, ``Robust object re-identification with coupled
  noisy labels,'' \emph{International Journal of Computer Vision}, vol. 132,
  no.~7, pp. 2511--2529, 2024.

\bibitem{tan2021co}
C.~Tan, J.~Xia, L.~Wu, and S.~Z. Li, ``Co-learning: Learning from noisy labels
  with self-supervision,'' in \emph{Proceedings of the 29th ACM international
  conference on multimedia}, 2021, pp. 1405--1413.

\bibitem{kim2024learning}
S.~Kim, D.~Lee, S.~Kang, S.~Chae, S.~Jang, and H.~Yu, ``Learning discriminative
  dynamics with label corruption for noisy label detection,'' in
  \emph{Proceedings of the IEEE/CVF Conference on Computer Vision and Pattern
  Recognition}, 2024, pp. 22\,477--22\,487.

\bibitem{huang2019o2u}
J.~Huang, L.~Qu, R.~Jia, and B.~Zhao, ``O2u-net: A simple noisy label detection
  approach for deep neural networks,'' in \emph{Proceedings of the IEEE/CVF
  international conference on computer vision}, 2019, pp. 3326--3334.

\bibitem{rolnick2017deep}
D.~Rolnick, A.~Veit, S.~Belongie, and N.~Shavit, ``Deep learning is robust to
  massive label noise,'' \emph{arXiv preprint arXiv:1705.10694}, 2017.

\bibitem{winn2005locus}
J.~Winn and N.~Jojic, ``Locus: Learning object classes with unsupervised
  segmentation,'' in \emph{Tenth IEEE International Conference on Computer
  Vision (ICCV'05) Volume 1}, vol.~1.\hskip 1em plus 0.5em minus 0.4em\relax
  IEEE, 2005, pp. 756--763.

\bibitem{wang2023metaviewer}
R.~Wang, H.~Sun, Y.~Ma, X.~Xi, and Y.~Yin, ``Metaviewer: Towards a unified
  multi-view representation,'' in \emph{Proceedings of the IEEE/CVF Conference
  on Computer Vision and Pattern Recognition}, 2023, pp. 11\,590--11\,599.

\bibitem{yang2010bag}
Y.~Yang and S.~Newsam, ``Bag-of-visual-words and spatial extensions for
  land-use classification,'' in \emph{Proceedings of the 18th SIGSPATIAL
  International Conference on Advances in Geographic Information Systems},
  2010, pp. 270--279.

\bibitem{lin2022dual}
Y.~Lin, Y.~Gou, X.~Liu, J.~Bai, J.~Lv, and X.~Peng, ``Dual contrastive
  prediction for incomplete multi-view representation learning,'' \emph{IEEE
  Transactions on Pattern Analysis and Machine Intelligence}, vol.~45, no.~4,
  pp. 4447--4461, 2022.

\bibitem{zhang2023provable}
Q.~Zhang, H.~Wu, C.~Zhang, Q.~Hu, H.~Fu, J.~T. Zhou, and X.~Peng, ``Provable
  dynamic fusion for low-quality multimodal data,'' in \emph{International
  Conference on Machine Learning}.\hskip 1em plus 0.5em minus 0.4em\relax PMLR,
  2023, pp. 41\,753--41\,769.

\bibitem{geng2021uncertainty}
Y.~Geng, Z.~Han, C.~Zhang, and Q.~Hu, ``Uncertainty-aware multi-view
  representation learning,'' in \emph{Proceedings of the AAAI Conference on
  Artificial Intelligence}, vol.~35, no.~9, 2021, pp. 7545--7553.

\bibitem{huang2025trusted}
H.~Huang, C.~Qin, Z.~Liu, K.~Ma, J.~Chen, H.~Fang, C.~Ban, H.~Sun, and Z.~He,
  ``Trusted unified feature-neighborhood dynamics for multi-view
  classification,'' in \emph{Proceedings of the AAAI conference on artificial
  intelligence}, vol.~39, no.~16, 2025, pp. 17\,413--17\,421.

\bibitem{xu2025beyond}
C.~Xu, Z.~Wen, J.~Zhao, W.~Zhao, J.~Yu, H.~Chen, Z.~Guan, and W.~Zhao, ``Beyond
  equal views: Strength-adaptive evidential multi-view learning,'' in
  \emph{Proceedings of the 33rd ACM International Conference on Multimedia},
  2025, pp. 1278--1287.

\bibitem{duandeep}
S.~Duan, Y.~Sun, D.~Peng, G.~Duan, X.~Peng, and P.~Hu, ``Deep fuzzy multi-view
  learning for reliable classification,'' in \emph{Forty-second International
  Conference on Machine Learning}, 2025.

\end{thebibliography}

\vfill


\end{document}